%% file: ggnn_arxiv.tex
\documentclass[10pt,final,journal,compsoc]{IEEEtran}

\ifCLASSOPTIONdraftcls
\else
\fi

\ifCLASSOPTIONfinal
\else
\fi

\ifCLASSOPTIONcompsoc
  \usepackage[nocompress]{cite}
\else
  \usepackage{cite}
\fi

\ifCLASSINFOpdf
\else
\fi

\usepackage{times}
\usepackage{epsfig}
\usepackage{graphicx}
\usepackage{amsmath}
\usepackage{amssymb}
\usepackage{color}
\usepackage{xcolor}
\usepackage{booktabs}
\usepackage{multirow}
\usepackage{subcaption}
\usepackage{siunitx}
\usepackage[algoruled]{algorithm2e}
\usepackage[font=small,labelfont=bf,tableposition=top]{caption}

\usepackage{xpunctuate}

\input{macros.tex}

\usepackage[pagebackref=true,breaklinks=true,letterpaper=true,colorlinks,bookmarks=false]{hyperref}

\usepackage{etoolbox}
\robustify\bfseries

\begin{document}
\title{GGNN: Graph-based GPU Nearest Neighbor Search}

\author{Fabian~Groh,
        Lukas~Ruppert, %
        Patrick~Wieschollek, %
        and Hendrik~P.A.~Lensch%
\IEEEcompsocitemizethanks{\IEEEcompsocthanksitem During the work of this paper, all authors were member of the Computer Graphics group at the University of Tübingen, Germany.
\IEEEcompsocthanksitem Patrick Wieschollek and Fabian Groh are now with Amazon, Tübingen, Germany. This work has been done prior to joining Amazon.
\IEEEcompsocthanksitem This is the author's version of the paper. The final published version can be found here: \href{https://doi.org/10.1109/TBDATA.2022.3161156}{https://doi.org/10.1109/TBDATA.2022.3161156}
\IEEEcompsocthanksitem \copyright 2022 IEEE.  Personal use of this material is permitted.  Permission from IEEE must be obtained for all other uses, in any current or future media, including reprinting/republishing this material for advertising or promotional purposes, creating new collective works, for resale or redistribution to servers or lists, or reuse of any copyrighted component of this work in other works.
}}

\IEEEtitleabstractindextext{%
\begin{abstract}
  \input{tex/abstract.tex}
\end{abstract}

\begin{IEEEkeywords}
    Nearest neighbor searches, Graph and tree search strategies, Information retrieval, Approximate search, Similarity search, Big data.
\end{IEEEkeywords}}

\maketitle

\begin{figure*}
  \centerline{
    \includegraphics{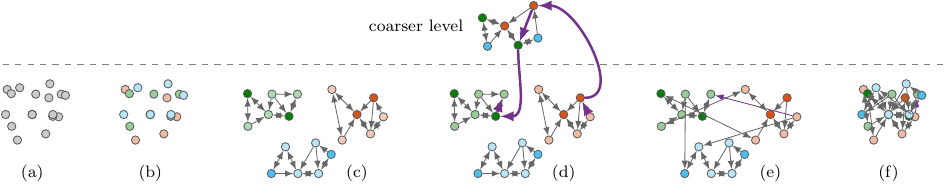}}
    \caption{
    Illustration of the bottom-up construction of the kNN-graph for a dataset (a). After random partitioning (b), a kNN-graph is constructed for each partition (c). A selection of nodes builds a coarser kNN-graph (d), which is used to propagate links between partitions. These links are used to merge several partitions (e) into a final kNN-graph (f).}
    \label{fig:merging_new}
\end{figure*}

\IEEEdisplaynontitleabstractindextext

\IEEEpeerreviewmaketitle

\input{tex/defines.tex}

\input{tex/introduction.tex}
\input{tex/relatedwork.tex}
\input{tex/background.tex}
\input{tex/method_new.tex}
\input{tex/evaluation.tex}
\input{tex/conclusion.tex}

\ifCLASSOPTIONdraftcls
\appendices
\input{tex/appendix.tex}
\fi

\ifCLASSOPTIONcompsoc
  \section*{Acknowledgments}
\else
  \section*{Acknowledgment}
\fi

This work has been partially funded by the Deutsche Forschungsgemeinschaft (DFG, German Research Foundation) under Germany’s Excellence Strategy
-- EXC number 2064/1 -- Project number 390727645 and SFB 1233, TP 02 -- Project number 276693517.
It was supported by the German Federal Ministry of Education and Research (BMBF): Tübingen AI Center, FKZ: 01IS18039A.

\appendices

\input{tex/appendix.tex}
\ifCLASSOPTIONcaptionsoff
  \newpage
\fi

\bibliographystyle{IEEEtran}
\bibliography{ggnn_arxiv}

\begin{IEEEbiography}[{\includegraphics[width=1in,height=1.25in,clip,keepaspectratio]{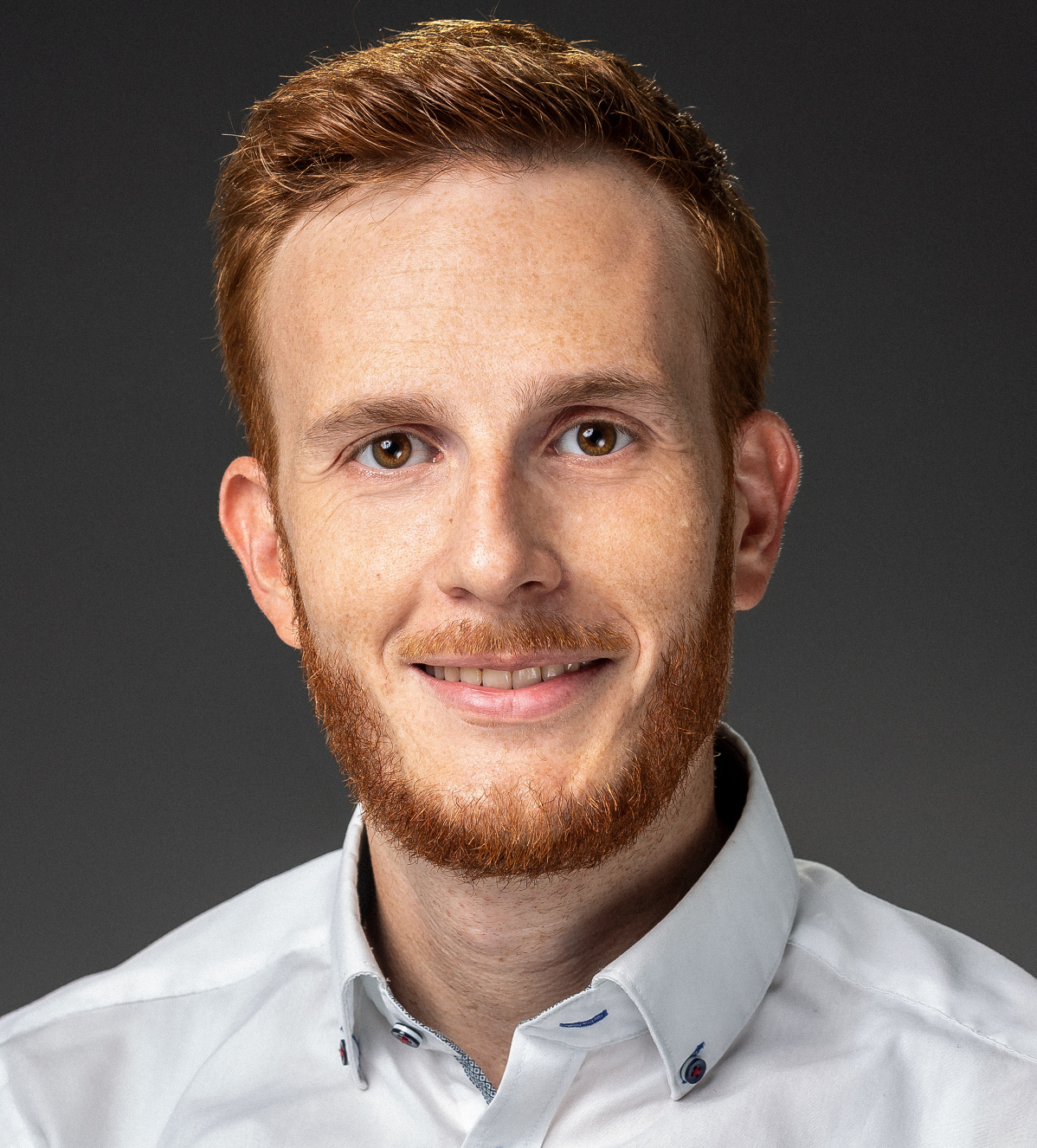}}]{Fabian Groh}
received his diploma in computer science in 2013 from Ulm University, Germany. He is working towards a PhD degree at the computer graphics group of the University of Tübingen. His primary research topics are processing and learning on unstructured data in large-scale settings by utilizing GPU-based designs. In 2020 he joined Amazon Tübingen.
\end{IEEEbiography}

\begin{IEEEbiography}[{\includegraphics[width=1in,height=1.25in,clip,keepaspectratio]{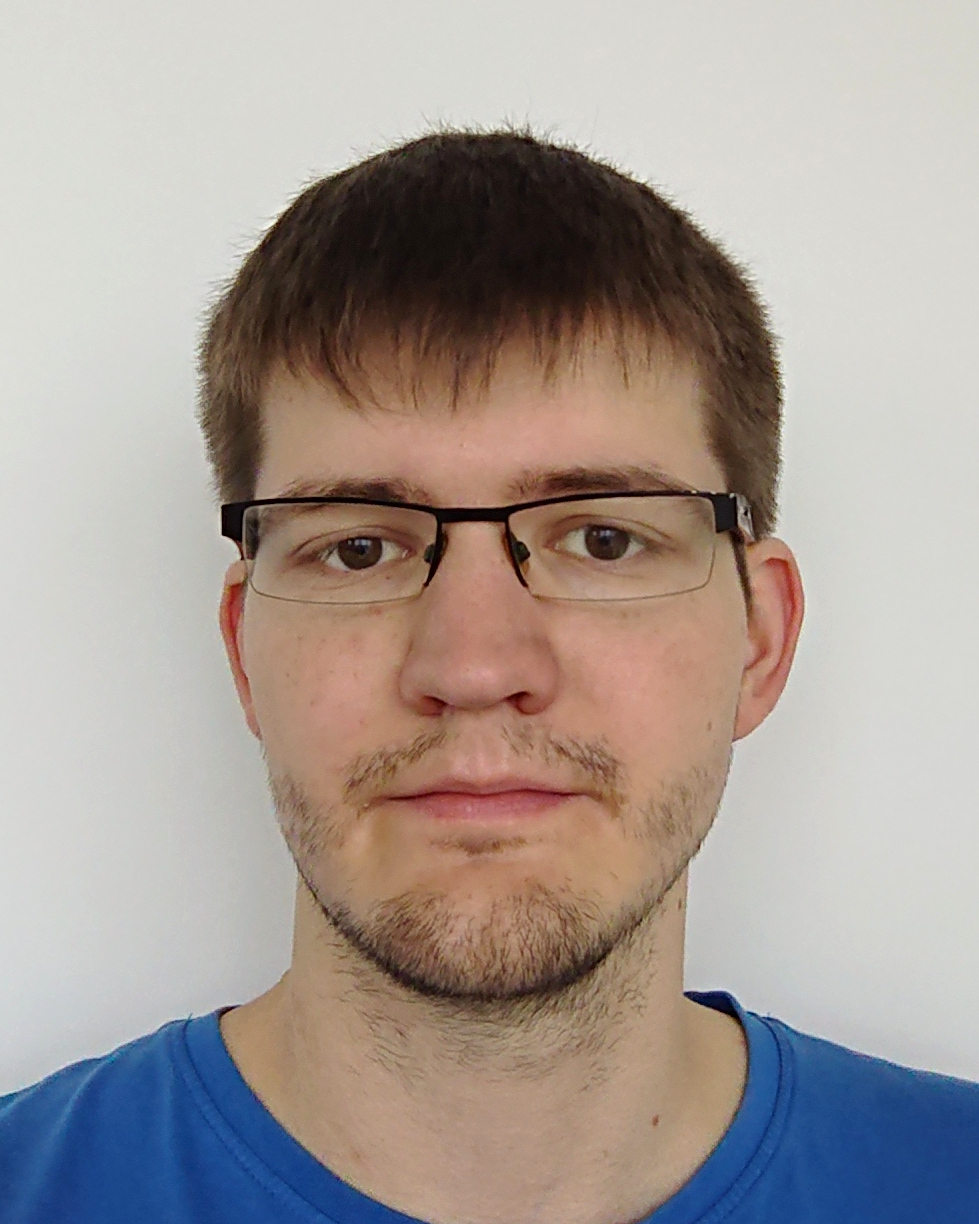}}]{Lukas Ruppert}
received his master of science in computer science in 2019 from University of Tübingen.
He is working on his Ph.D. in computer graphics with a focus on physically based rendering.
\end{IEEEbiography}

\begin{IEEEbiography}[{\includegraphics[width=1in,height=1.25in,clip,keepaspectratio]{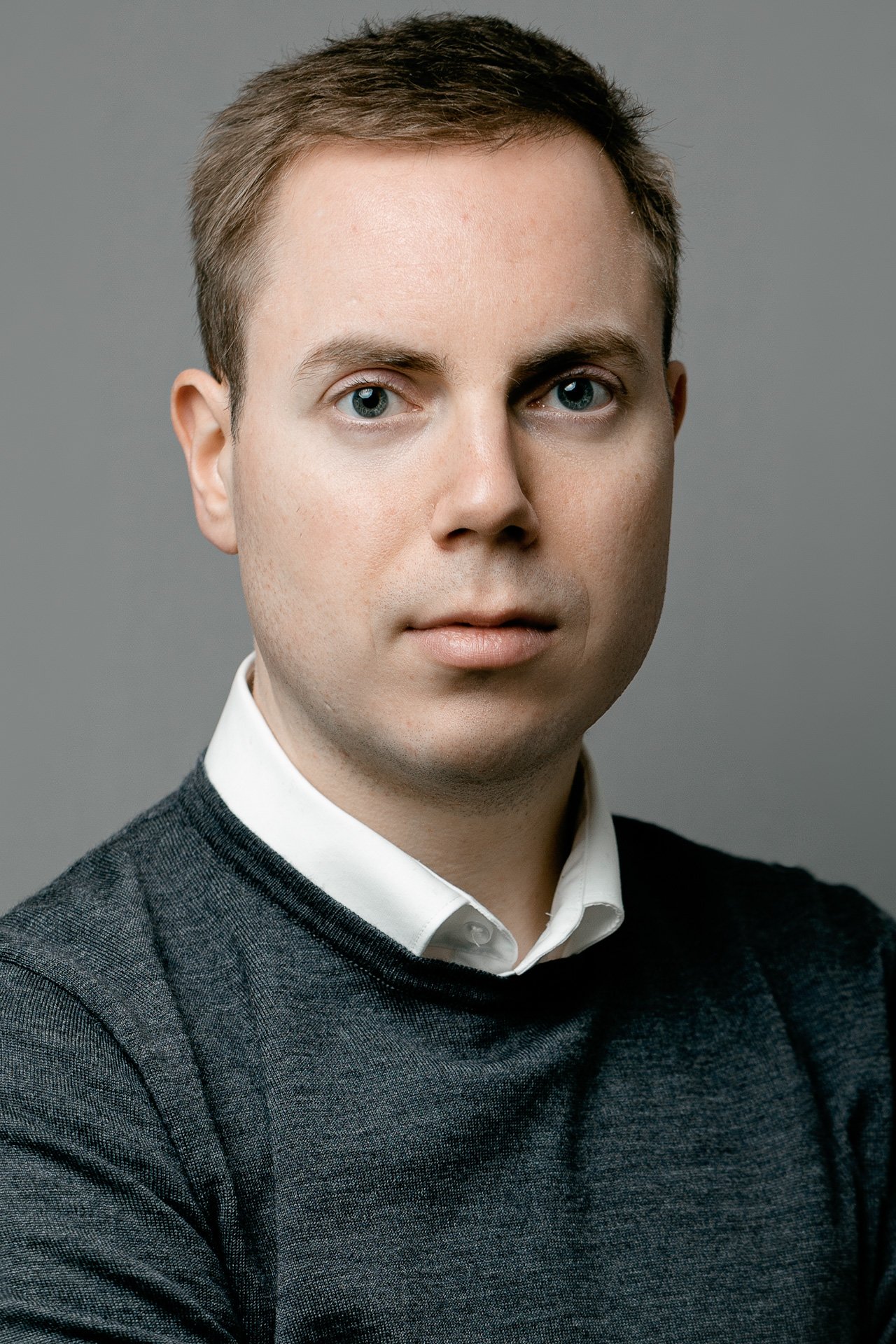}}]{Patrick Wieschollek}
received his master of science in 2014 from Friedrich-Schiller University, Germany. He wrote his PhD thesis at the computer graphics group of the University of Tübingen and Max Planck Institute for Intelligent Systems in Tübingen. His primary research topic is developing data-driven approaches for multi-frame and unstructured data in large-scale settings. In 2019 he joined Amazon Tübingen.
\end{IEEEbiography}

\begin{IEEEbiography}[{\includegraphics[width=1in,height=1.25in,clip,keepaspectratio]{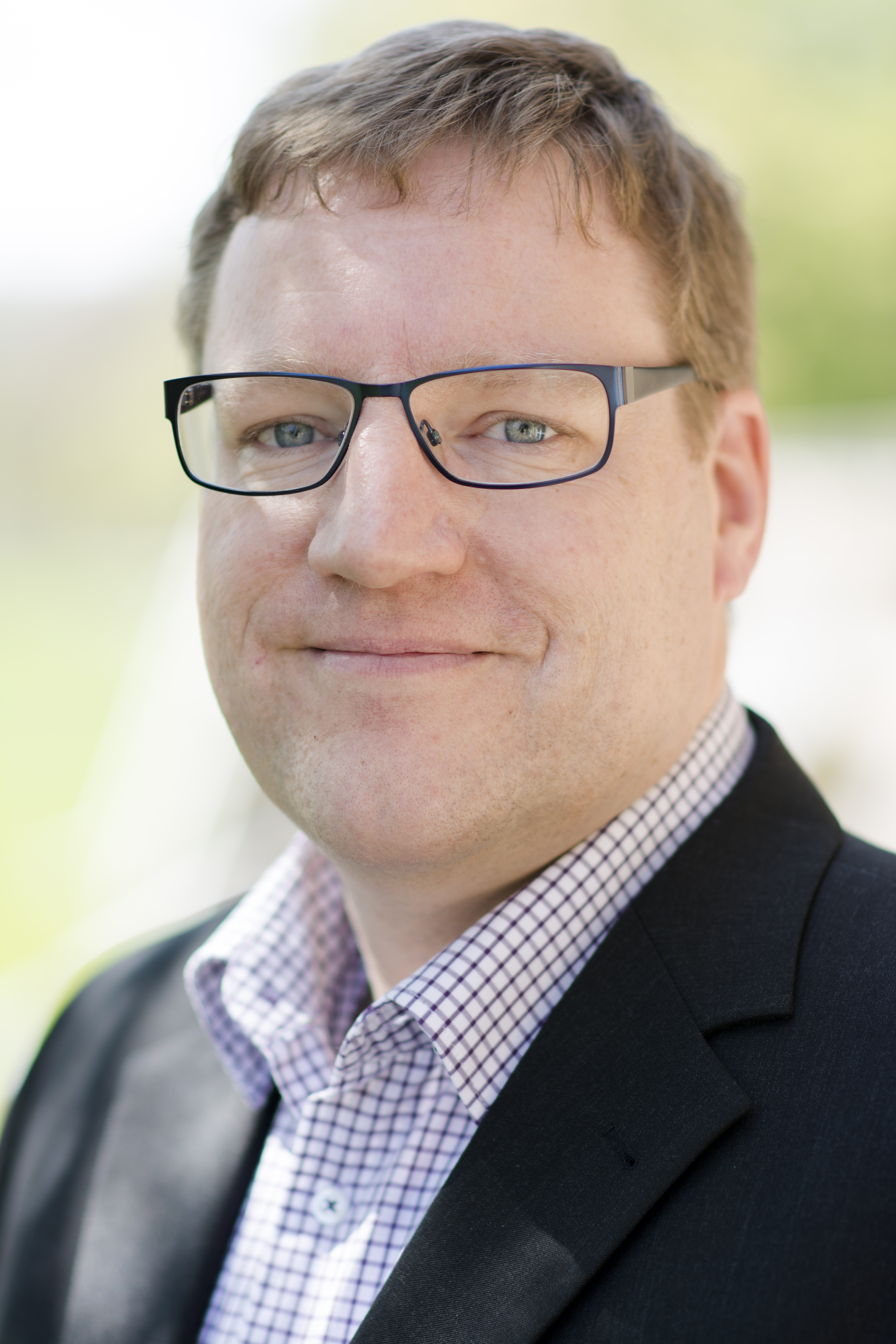}}]{Hendrik Lensch}
Hendrik P. A. Lensch holds the chair for computer graphics at Tübingen University. He received his diploma in computers science from the University of Erlangen in 1999. He worked as a research associate at the computer graphics group at the Max-Planck-Institut für Informatik in Saarbrücken, Germany, and received his PhD from Saarland University in 2003. Hendrik Lensch spent two years (2004-2006) as a visiting assistant professor at Stanford University, USA. From 2009 to 2011 he was a full professor at the Institute for Media Informatics at Ulm University, Germany. In his career, he received the Eurographics Young Researcher Award 2005 and was awarded an Emmy Noether Fellowship by the German Research Foundation (DFG) in 2007. His research interests include 3D appearance acquisition, computational photography, machine learning, global illumination and image-based rendering, and massively parallel programming.

\end{IEEEbiography}

\end{document}

%% file: macros.tex
\newcommand{\citet} [1] {\cite{#1}}

\DeclareRobustCommand\eg{\xcommaafter{\xperiodafter{\emph{e.g}}}}

\DeclareRobustCommand\ie{\xcommaafter{\xperiodafter{\emph{i.e}}}}

\DeclareRobustCommand\wrt{\xperiodafter{\emph{w.r.t}}}

\DeclareRobustCommand\etal{\xperiodafter{\emph{et al}}}

\newcommand{\norm}[1]{\left\Vert #1 \right\Vert}

\DeclareMathOperator*{\argmin}{arg\,min}

%% file: tex/abstract.tex
Approximate nearest neighbor (ANN) search in high dimensions is an integral part of several computer vision systems and gains importance in deep learning with explicit memory representations. %
Since PQT~\cite{wieschollek2016efficient}, FAISS~\cite{johnson2019billion}, and SONG~\cite{zhao2020song} started to leverage the massive parallelism offered by GPUs,
GPU-based implementations are a crucial resource for today's state-of-the-art ANN methods. %
While most of these methods allow for faster queries, less emphasis is devoted to accelerating the construction of the underlying index structures. %
In this paper, we propose a novel GPU-friendly search structure based on nearest neighbor graphs and information propagation on graphs. %
Our method is designed to take advantage of GPU architectures to accelerate the hierarchical construction of the index structure and for performing the query. %
Empirical evaluation shows that GGNN significantly surpasses the state-of-the-art CPU- and GPU-based systems in terms of build-time, accuracy and search speed.

%% file: tex/defines.tex
\newcommand{\Kn}{$k_\text{nn}$}
\newcommand{\Ks}{$k_\text{sym}$}

\SetStartEndCondition{ }{}{}%
\SetKwProg{Fn}{Function}{:}{end}
\SetKwFunction{BT}{BuildGraph}%
\SetKwFunction{Init}{init\_tree}%
\SetKwFunction{Merge}{merge}%
\SetKwFunction{Sym}{sym}%
\SetKwFunction{Query}{query}%
\SetKwFunction{Fetch}{fetch}%
\SetKwFunction{SymQuery}{sym\_links}%
\SetKwFunction{Stats}{stats}%
\SetKwFunction{Select}{select}%
\SetKwFunction{Refine}{RefineGraph}

\SetKw{KwTo}{ to }
\SetKwFor{For}{for}{\string:}{}%
\SetKwFor{ForEach}{for each}{\string:}{}%
\SetKwFor{ParFor}{parfor}{\string:}{}%
\SetKwIF{If}{ElseIf}{Else}{if}{:}{elif}{else:}{}%
\SetKwFor{While}{while}{:}{fintq}%
\SetKw{Break}{break}
\AlgoDontDisplayBlockMarkers
\SetAlgoNoEnd
\SetAlgoNoLine%
\DontPrintSemicolon
\newcommand\commentfont[1]{\relsize{-1}\ttfamily{#1}}
\SetCommentSty{commentfont}

\SetKwData{LT}{$l_{top}$}
\SetKwData{LB}{$l_{m}$}
\SetKwData{Layer}{$i$}
\SetKwData{L}{$L$}
\SetKwData{Li}{$l_i$}
\SetKwData{K}{$k$}
\SetKwData{KL}{$k_{nn}$}
\SetKwData{KF}{$k_{sym}$}
\SetKwData{XIb}{$\tau_{build}$}
\SetKwData{XIq}{$\tau_{q}$}
\SetKwData{XIg}{$\tau$}
\SetKwData{Base}{$\mathcal{X}$}
\SetKwData{Tree}{\texttt{graph}}
\SetKwData{Buffer}{$b$}

\SetKwData{N}{$z$}
\SetKwData{Q}{$q$}
\SetKwData{Z}{$z$}
\SetKwData{X}{$x$}
\SetKwData{C}{$c$}
\SetKwData{D}{$d$}
\SetKwData{Xhat}{$\hat{x}$}
\SetKwData{Xhati}{$\hat{x}_i$}
\SetKwData{G}{$G$}
\SetKwData{Sbf}{$\mathbf{s}$}
\SetKwData{S}{$s$}
\SetKwData{Si}{$\S_{i}$}
\SetKwData{Pbf}{$\mathbf{p}$}
\SetKwData{P}{$p$}
\SetKwData{A}{$a$}
\SetKwData{Cache}{$cache$}
\SetKwData{Best}{$best$}
\SetKwData{PrioQ}{$prioq$}
\SetKwData{Visited}{$visited$}
\SetKwData{Pc}{$p_{c_i}$}
\SetKwData{Dc}{$d_{c_i}$}
\SetKwData{Pcn}{$\underline{p_{c_{i+1}}}$}
\SetKwData{Dcn}{$\underline{d_{c_{i+1}}}$}
\SetKwData{Dcp}{$\underline{d_{c_{i-1}}}$}
\SetKwData{Empty}{$empty$}

\SetKwFunction{TopSeg}{getTopSegment}%
\SetKwFunction{StartPoint}{getStartPoint}%
\SetKwFunction{Fetch}{fetch}%
\SetKwFunction{Init}{init}%
\SetKwFunction{SetStart}{set\_start}%
\SetKwFunction{Add}{add}%
\SetKwFunction{Push}{push}%
\SetKwFunction{Pop}{pop}%
\SetKwFunction{Dist}{dist}%
\SetKwFunction{Criteria}{criteria}%
\SetKwFunction{Connected}{is\_connected}%
\SetKwFunction{Transform}{transform}%
\SetKwFunction{Clear}{clear\_visited}%
\SetKwFunction{Skip}{skip\_criteria}%
\SetKwFunction{SkipSym}{skip\_criteria\_sym}%
\SetKwFunction{Tlayer}{layer}%
\SetKwFunction{Begin}{is\_begin}%
\SetKwFunction{NEnd}{is\_not\_end}%
\SetKwFunction{MoveHead}{move\_head\_forward}%

%% file: tex/introduction.tex
\section{Introduction}
Approximate nearest neighbor (ANN) search plays a crucial and long-standing role in various domains, including databases, computer vision, autonomous vehicles, personalized medicine, and machine learning.  %
Since collecting large amounts of data became easier, %
the creation of a scalable and efficient data structure for retrieving similar items has become an active research topic. %
Despite all recent advances, the only available method for guaranteed retrieval of the exact nearest neighbor in high dimensions is still exhaustive search, due to the curse of dimensionality~\cite{Weber1998AQA}. %
Even when leveraging modern hardware, it remains impractical to perform an exhaustive search over billions of high-dimensional data points. %
Instead, most popular methods relax the problem by searching for an entry that is likely to be the nearest neighbor, accepting a minimal loss in accuracy. %

Besides designing a very fast GPU-based approximate kNN query algorithm, we address the efficient construction of a hierarchical graph-based search structure. %
Building time becomes more and more important in dynamically growing or changing datasets for on-the-fly analysis,
e.g. in correspondence and feature matching for tracking and object recognition in videos,
for newly embedded vectors of neural networks similar to DEEP1B \cite{Babenko2016EfficientIO}, or in recommender systems \cite{bobadilla2013recommender}. %
Our graph-construction explicitly determines the true k nearest neighbors of each point in the dataset with high probability.
Solving this for-all task is highly relevant in $n$-body problems like salient point estimation, kernel-density computation, cluster analysis, retrieval of more robust prototypes, or feature matching in embeddings. %

To keep up with the scale of data that is produced day by day, modern approaches use index structures that are heavily tailored towards exploiting the massive parallelism of GPUs~\cite{wieschollek2016efficient,johnson2019billion,chen2019vlq,harwood2016fanng} or custom hardware~\cite{PQFPGA},
as opposed to previous CPU-based methods~\cite{flann_pami_2014,PQ,1BPQ,localPQ,optimizedPQ,invertedmultiindex,Babenko_2015_CVPR,Babenko_2014_CVPR}.

The quality of the recall heavily depends both on the choice of the search structure and the executed search (query) itself. %
Structures based on quantization or hashing/binning schemes~\cite{PQ,1BPQ,localPQ,optimizedPQ,invertedmultiindex,Babenko_2015_CVPR,Babenko_2014_CVPR,wieschollek2016efficient,chen2019robustiq,johnson2019billion,Matsui2015PQTableFE} can be built efficiently, but typically suffer from relatively low recall rates as enumerating and visiting neighboring cells is exhaustive in high dimensions. %
Better recall rates are recently achieved by graph-based methods~\cite{malkov2018efficient,li2019approximate,harwood2016fanng,dong2011efficient,chen2009fast,Warashina2014EfficientKN,fu2016efanna,fu2019fast,jieren2020hm-ann}. %
Existing methods for constructing effective search-graphs, e.g.~\cite{malkov2018efficient}, update varying-sized edge lists sequentially. %
They are highly dependent on global memory synchronization and difficult to parallelize effectively beyond a few cores. %
Hence, their construction times are not scaling well and are measured in hours or even days~\cite{Matsui2015PQTableFE,jieren2020hm-ann}.

Given a precomputed graph structure, a query traverses the edges of the graph to decrease the distance to a query point. %
It needs to compute the distances to all neighbors of the currently investigated node, move to the next-best point and store which points have already been visited.
All decisions are made locally and independent for each query, rendering the query algorithm an ideal candidate for parallelization.
However, a parallel implementation needs to carefully address a number of issues to be efficient. %
For one, the available memory bandwidth for loading each vector to compare with might become a limiting factor. %
Here, Optane memory~\cite{jieren2020hm-ann} or GPUs~\cite{zhao2020song} offer a solution. %
Secondly, on GPUs one needs to cope with the limited amount of available per-thread or per-block memory when storing the list of visited points.

Hence, we propose a GPU-friendly query design on thread-block level with high on-chip resource utilization through a fully parallel multi-purpose cache and a fixed number of neighbors per point. %
Further, we introduce a novel technique for very fast graph construction that utilizes the fast parallel query algorithm to iteratively merge multiple sub-graphs together. %
This bottom-up construction scheme is sketched in Figure~\ref{fig:merging_new}. %
Hierarchical graphs are used as a global optimization substitution to overcome gaps in local connectivity, in particular during construction. %
Additionally, our local symmetric linking approach reduces the amount of redundant links in the graph to circumvent negative impacts of overflows in the memory-limited caches. %

Our approach is also inherently suited for batch processing by already generating multiple sub-graphs at the same time. %
By forgoing the final merge of several sub-graphs into one combined graph, one can easily obtain independent yet effective sub-graphs for parts of large-scale datasets which would otherwise not fit in memory.
These can easily be processed even on multiple GPUs at the same time where each query then needs to be executed once on each shard.
This simple yet effective method allows for optimal multi GPU utilization. %

To summarize our main contributions, we propose a method for extremely fast approximate kNN-search on GPUs for high-dimensional data. %
We focus not only on fast query time, but also on a very efficient construction of the index structure, while still achieving high recall rates of the true nearest neighbours.

As evidenced by our empirical evaluation, the presented scheme outperforms existing approaches concerning both the construction as well as the query time. %
At the same time, the recall rate is consistently high and can be traded in for even faster construction or query time. %
We present a multi-GPU scheme that is capable of achieving above 99\% recall even for common benchmark datasets with billions of high-dimensional entries. %

%% file: tex/relatedwork.tex
\section{Related Work}
A large amount of literature exists on designing structures that accelerate a nearest neighbor search. Besides traditional approaches~\cite{KD-tree, flann_pami_2014}, most popular techniques rely either on data quantization in clusters~\cite{PQ,1BPQ,localPQ,optimizedPQ,invertedmultiindex,Babenko_2015_CVPR,Babenko_2014_CVPR,wieschollek2016efficient,chen2019robustiq,johnson2019billion,Matsui2015PQTableFE} or building neighborhood graphs~\cite{harwood2016fanng,dong2011efficient,chen2009fast,Warashina2014EfficientKN,fu2016efanna,fu2019fast,jieren2020hm-ann}. %
To achieve peak performance, most of these methods compute a compressed representation for each entry as large datasets will not fit into fast memory. %
There exist several strategies to compute such a compression. While hashing methods~\cite{LSH,Andoni2008,Korman2011} produce compact binary codes, quantization-based methods reuse centroids by assigning each data point a unique identifier based on the centroid to which they belong. %
It has been empirically shown that quantization methods are more accurate than various hashing methods~\cite{PQ,flann_pami_2014}.

\textit{Quantization methods} for nearest neighbor search using clustering methods were popularized by J{\'e}gou~\etal~\cite{PQ} while originally being introduced in~\cite{originalPQ}. %
Such index structures, like IVFADC~\cite{PQ}, partition the high-dimensional search space into disjoint Voronoi cells described by a set of centroids obtained by Vector Quantization (VQ)~\cite{VQ}. %
The idea has been extended later by Babenko~\etal~\cite{invertedmultiindex}, where the high-dimensional vector-space is factored in orthogonal subspaces. %
Hereby, each vector is assigned to a centroid independently for each subspace according to a separate codebook that resides there. %
Wieschollek~\etal~\cite{wieschollek2016efficient} proposed a hierarchical representation of the codebook besides demonstrating superior performance using a GPU. %
Johnson~\etal~\cite{johnson2019billion} ported IVFADC~\cite{PQ} to the GPU in combination with a fast GPU-based implementation for k-selection, \ie returning the $k$ lowest-valued elements from a given list (a crucial part of quantization based methods). %
They are the first employing multi-GPU parallelism by replication and sharding. Their work forms the library ``FAISS''. %
Eventually, Chen~\etal~\cite{chen2019robustiq} proposed a GPU based method RobustiQ overcoming the memory limitations of FAISS by extending the idea of Line Quantization from ~\cite{wieschollek2016efficient} in a hierarchical fashion. Still, the reported distances are only an approximation of the true distance. %

All hashing and quantization-based indexing schemes share the same problem that they partition the space into cells.
While the containing cell for a query might be found very efficiently, the exact nearest neighbor might be across the boundary to one of the neighboring cells. Determining and visiting all neighboring cells in high dimension is a problem severely limiting these approaches.

\textit{kNN-graph} based methods are another way to accelerate the query process.
Our presented approach belongs to this category. %
The main idea is to link each point from the search space to $k$ of its nearby points. %
Each query will start at a random guess in the dataset. Then, the guess itself is refined by replacing it with a better point from the $k$ linked neighbor points. %
Chen~\etal~\cite{chen2009fast} propose a fast divide and conquer strategy for computing such kNN-graphs. %
Done~\etal~\cite{dong2011efficient} introduced NN-descent for using kNN-graphs to accelerate NN-search. %
Hereby, each point maintains a list of its own nearest neighbors and points where itself is considered as a nearest neighbor. %
This has been later extended~\cite{Warashina2014EfficientKN} to make use of MapReduce. %
EFANNA as a multiple hierarchical index structure uses a truncated KD-tree to build a kNN-graph~\cite{fu2016efanna}. %

In the ideal case, a kNN-graph augmented with additional links could guarantee that for an arbitrary starting point the NN-descent will converge to the correct solution. %
Computing such a graph with additional links at scale is not practicable. Therefore, several methods exist to at least approximate such a graph~\cite{fu2019fast,harwood2016fanng}. %
Fu~\etal~\cite{fu2019fast} introduce NSG as an approximation.
To reduce the overall number of edges, their optimization tries to lower the out-degree individually per node.  %
Their method can scale beyond multiple cores, outperforming a GPU approach~\cite{johnson2019billion} on a benchmark dataset.
Harwood~\etal~\cite{harwood2016fanng} suggest an alternative approach for constructing such a graph. Starting from a fairly dense graph, they remove ``shadowed edges'', which are redundant when considering traversing paths during a query. They showed promising results using the GPU but only on rather small datasets as their build time is rather high.
Malkov~\etal~\cite{malkov2018efficient} construct a hierarchical graph structure to accelerate the nearest neighbor search.

For graph-based methods, memory throughput is often the limiting factor.
Ren~\etal~\cite{jieren2020hm-ann} propose an optimized construction of HNSW-style search graphs on out-of-core datasets using heterogeneous memory.
Using fast Optane memory allows them to quickly query even billion-scale datasets at high accuracy.
Zhao~\etal~\cite{zhao2020song} propose GPU-based "search on graph" (SONG) using index structures built using either HNSW~\cite{malkov2018efficient} or NSG~\cite{fu2019fast},
achieving significant speedup over CPU-based queries in most cases.

Our method is most similar to HNSW~\cite{malkov2018efficient} with its hierarchical construction,
but it is very carefully tuned for optimal parallelism of construction tasks and also differs during the query.

%% file: tex/background.tex
\section{Background}
\label{sec:background}
In this section, we formally introduce the approximate nearest neighbor (ANN) problem statement and used notation.

\subsection{Nearest Neighbor Search}
\label{subsec:annsearch}
The nearest neighbor problem retrieves a point $x^\star$ from a dataset $\mathcal{X}=\left\{x_1,\ldots,x_n\right\}$ that has the smallest distance to a query $q$.
For the sake of simplicity, here, we assume an Euclidean space (${\mathcal{X}\subset\mathbb{R}^d}, {q\in\mathbb{R}^d}$) and Euclidean distances ($\norm{\cdot}_2$). %
The nearest neighbor $x^\star\in\mathcal{X}$ of $q$ therefore is defined as
\begin{align}
    x^\star = \argmin_{x\in \mathcal{X}} \norm{q-x}_2. \label{eq:nnsearch}
\end{align}
Similarly, the $k$-nearest neighbor search retrieves the $k$ closest entries from $\mathcal{X}$ for a given query.
As finding the exact nearest neighbor might be costly, we may accept points in $\mathcal{X}$ which are close to $q$ and therefore deliver an approximate solution to Eq.~\eqref{eq:nnsearch}.

\subsection{KNN Graph}
\label{subsec:knngraph}
In a kNN-graph, each point $x$ form the dataset $\mathcal{X}$ represents one node in the graph $G$.
Further, we define $\mathcal{N}_x\subseteq \mathcal{X}$ as a local neighborhood of $x$ with $k$ elements and defer the details on how to construct $\mathcal{N}_x$ to the next section.
The edges $E$ of the graph are then defined as $(x,y)$ where $y \in \mathcal{N}_x$.
Note that the resulting graph is a directed graph $G=(\mathcal{X}, E)$, where $E=\left\{(x,y)\;|\;x\in\mathcal{X}, y \in \mathcal{N}_x\right\}$
and $(x,y) \in E \nRightarrow (y,x) \in E$.

One greedy algorithm to find the nearest neighbor for a query point $q$ is \textit{NN-descent}~\cite{dong2011efficient}.
Starting from an initial guess ${x\in\mathcal{X}}$, the distance between $q$ and each neighboring point $y\in \mathcal{N}_x$ is computed.
If any $y\in \mathcal{N}_x$ is closer to $q$ than $x$, the guess $x$ is replaced by the closest point from $\mathcal{N}_x$ to $q$.
This process iterates until no point in $\mathcal{N}_x$ has a smaller distance to $q$ than $x$.
However, as the current $x$ might not provide an edge into the right search direction, this greedy algorithm might get stuck in a local minimum on a pure kNN-graph.

\subsubsection{Common Pitfalls}
Since the NN-descent is a greedy search, it offers no guarantee on finding the exact solution. Some of the reasons are listed below:

\textit{Connectivity:}
As a kNN-graph is a directed graph, $y$ might be directly connected to $x$, being its nearest neighbor, hence ${y\in\mathcal{N}_x}$.
But this does not imply that the inverse link exists (${y\in\mathcal{N}_x \nRightarrow x\in\mathcal{N}_y}$).
Therefore, the construction of an augmented (diversified) kNN search-graph has to deal with synchronizing outgoing and incoming (inverse) edges.

\textit{Gaps in high-dimensional spaces:} As each point is only linked to a finite number of local neighbors, there exist pathological cases (even in 2D), where close-by points are not directly connected at all.
Such a case is illustrated in Figure~\ref{fig:slack}.
Due to the gap, the true nearest neighbor will not be found.
Computing an idealized monotonic relative neighborhood graph~\cite{fu2019fast} (MRNG) would avoid this issue, but requires a strongly varying connectivity not suitable for parallel approaches -- besides its additional computational burden.

\textit{Degree of nodes:} There exists a trade-off when choosing the cardinality of $\mathcal{N}_{x}$ for any $x\in\mathcal{X}$.
Having fewer edges amplifies the previously described issues, but also reduces the number of necessary comparisons at each step.
Conversely, having many edges allows the greedy search to escape from local neighborhoods but increases the cost for each iteration.

%% file: tex/method_new.tex
\section{GPU-Based Nearest Neighbor Search}
Searching for the $k$ nearest neighbors for multiple queries, given some graph structure, is the central operation for various applications and also the main building block for our graph construction (see Section~\ref{sec:method}). %
Achieving high recall rates at very short times is the primary goal of our parallel GPU-based search-algorithm on kNN-graph structures. %
The algorithm can, within bounds, be tuned for quality or speed, offering a solution to a broad range of use-cases,
accommodating those requiring high recall rates as well as real-time algorithms with reduced quality constraints.

The main idea is to highly parallelize the search by utilizing one thread-block per query. %
In contrast to the na\"ive solution of a query per thread, the main advantage is to bundle enough on-chip resources that all meta-data can be kept on very fast memory. %
This is of utmost importance, since graph-based methods still need to load a large amount of neighborhood information as well as vectors from global memory for distance computations. %
In addition, the thread-block approach guarantees high memory bandwidth through very efficient coalesced memory accesses. %
It also allows to perform more computation in parallel
since common tasks like distance computations or the maintenance of priority queues and visited lists can be parallelized.
Furthermore, individual queries that vary in runtime can be scheduled independently.

\subsection{GPU-based Search with Backtracking}
\label{subsec:query}
Assuming a diversified (see Section~\ref{subsec:symmetry}) kNN-graph with given starting points $\Sbf\subset\mathcal{X}$, for a query $q\in \mathbb{R}^d$,
the simple greedy downhill search with backtracking from Algorithm~\ref{alg:query} is executed.
First, the cache structure (see Section~\ref{subsubsec:caching}) is initialized (\Init) with the query point and a slack factor $\tau$ (see Section~\ref{subsec:stopping}).
Then, the starting points \Sbf are introduced to the cache by \Fetch, which computes their distances to the query $q$ and adds them to the priority queue (\PrioQ).
As starting points \Sbf, we use the nodes on the top-layer of our hierarchical graph (see Section~\ref{subsec:querystart}).

Until there are no more unexplored points in the priority queue, all neighbors $\mathcal{N}_{\A}$ of the closest not yet visited point \A are being fetched to the cache. %
Essentially, performing a depth-first search in the graph.
Our cache structure manages a sorted list of the $k$ closest points (\Best) to the query which are observed during the traversal. %
Hence, the \Best list is returned as the result of the search.

\begin{algorithm}[ht]
    \caption{Basic query with backtracking.}
    \label{alg:query}
    \Fn{\Query{\G, \Sbf, \Q, \XIg, \K}}{
        \KwIn{search-graph \G with start~points~\Sbf, query point~\Q, slack factor~\XIg (see Section~\ref{subsec:stopping})}
        \KwOut{\K closest points to \Q}
        \Cache.\Init{\Q, \XIg} \tcc*[h]{(see Section~\ref{subsubsec:caching})}\;
        \Cache.\Fetch{\Sbf}\;
        \While{$(\A \leftarrow \Cache.\Pop{}) \neq \emptyset$}{
            \Cache.\Fetch{$\mathcal{N}_{\A}$} \tcc*[h]{(see Section~\ref{subsec:knngraph})}\;
        }
        \Return \Cache.\Best\;
    }{}
\end{algorithm}

\subsubsection{Caching on GPUs}
\label{subsubsec:caching}

The centerpiece of our fast GPU-based kNN search is a multi-purpose cache (see Algorithms~\ref{alg:query} and~\ref{alg:cache}). %
One of the major deficits of GPUs is their limited on-chip memory (\emph{register} and \emph{shared memory}\footnotemark[1]{}) that can be accessed with low latency. %

In particular, kNN-graph algorithms have a strong demand for highly dynamic structures like unpredictably growing lists of potential points to visit or of already visited points. %
These meta-data are perused and modified frequently during the query. %

\begin{figure}[ht]
    \centering
    \includegraphics[width=0.95\linewidth]{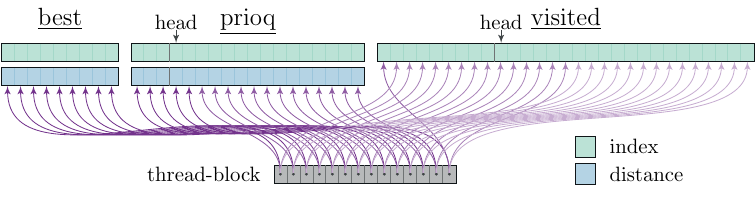}
    \caption{Our cache consists of a \protect\Best list, \protect\PrioQ (priority queue) and a \protect\Visited list.
    It resides entirely within shared memory.
    The \protect\Best list and \protect\PrioQ both contain indices and distances and are sorted by distance.
    The \protect\Visited list contains only indices to preserve memory.
    The \protect\PrioQ and the \protect\Visited list are implemented as ring buffers.
    This allows elements to be easily {\protect\Pop}ped from the front of the \protect\PrioQ and to overwrite old elements in the \protect\Visited list once it is full.
    The cache is accessed and maintained in parallel by the entire thread-block.}
    \label{fig:cache}
\end{figure}

The cache's memory layout is shown in Figure~\ref{fig:cache}. %
It consists of three major parts:
\begin{enumerate}
    \item \Best: A sorted list of the points closest to the query and their distances.
    \item \PrioQ: A priority-queue that manages points to be visited as a distance-sorted ring-buffer.
    \item \Visited: A ring-buffer that caches indices of already visited points in a first in first out (FIFO) fashion.
\end{enumerate}
All three parts are handled in \emph{shared memory}. %
The combined length is a multiple of the assigned threads, hence, every thread has multiple work items. %

The main methods of our cache are shown in Algorithm~\ref{alg:cache} and described in the following paragraphs. %
Note that for ease of readability synchronization barriers preventing race conditions are omitted. %
We refer to our published open source code\footnotemark[2]{} for more information. %
For the purpose of visualization, \emph{shared memory} variables are denoted with an underline, while normal variables can be assumed to live in \emph{register} space. %

\begin{algorithm}
    \caption{Cache Operations.}
    \label{alg:cache}
    \Fn{\Fetch{\Pbf}}{
        \KwIn{Point proposals \Pbf.}
        \ParFor{$\Pc \in \{\underline{\Best,\PrioQ,\Visited}\} \;$}{
            \For{$\P \in \underline{\Pbf} \;$}{
                \If{$\Pc = \P \;$}{
                    remove \P from \underline{\Pbf}\;
                }
            }
        }
        \For{$\P \in \underline{\Pbf} \;$}{
            \D $\leftarrow$ \Dist(\P) \tcc*[h]{Compute in parallel.}\;
            \If{$\Criteria(\D) \;$}{
                \Push(\P, \D)\;
            }
        }
    }{}
    \BlankLine
    \Fn(\tcc*[h]{Parallel insertion}){\Push{\P, \D}}{
        \KwIn{Point proposal \P with distance \D.}
        \ParFor{$\Pc, \Dc \in \{\underline{\Best,\PrioQ}\} \;$}{
            \tcc*[h]{Read \Pc and \Dc and sync.}\;
            \If{$\Dc \geq \D \;$}{
                \If{$ \NEnd(\C_i) \;$}{
                    \Pcn $\leftarrow$ \Pc\;
                    \Dcn $\leftarrow$ \Dc\;
                }
                \If{$\Begin(\C_i) \; \text{or} \; \Dcp < \D \;$}{
                    \underline{\Pc} $\leftarrow$ \P\;
                    \underline{\Dc} $\leftarrow$ \D\;
                }
            }
        }
    }{}
    \BlankLine
    \Fn(\tcc*[h]{Single-threaded routine}){\Pop{}}{
        \KwOut{Returns head of priority queue.}
        \P, \D $\leftarrow \underline{\PrioQ_{head}}$\;
        \If{not $\Criteria(\D)$}{
            \KwRet{\Empty}\;
        }
        \tcc*[h]{Remove point from \PrioQ.}\;
        $\underline{\PrioQ_{head}} \leftarrow \Empty$\;
        $\MoveHead(head_{\PrioQ})$\;
        \tcc*[h]{Add point to \Visited list.}\;
        $\underline{\Visited_{head}} \leftarrow \P$\;
        $\MoveHead(\Visited_{head})$\;
        \KwRet{\P}\;
    }{}

    \BlankLine
    \Fn{\Criteria{\D}}{
        \KwRet{$\D \le \D_{\Best_{\K}} + \xi $} \tcc*[h]{\!(see Section~\ref{subsec:stopping})\!}\;
    }{}
\end{algorithm}

\textbf{Fetch:}
Given a list of point proposals \Pbf, first, known elements are removed from the proposal list.
Each proposal is compared in parallel against the working items of each thread and removed in case of a match.
The remaining proposals are processed iteratively. %
The distance to the query point is computed in parallel (\Dist). %
Every thread reads its respective vector element(s) from memory and computes the element-wise result. %
A subsequent reduction produces the complete distance. %
For parallel primitives like reduction, we use NVIDIA's CUB\footnotemark[3]{} library for highly optimized implementations. %
The parallel coalesced loading and processing is highly effective in terms of GPU utilization. %
Points that are inside our \Criteria are {\Push}ed to the cache structure. %

\footnotetext[1]{\href{https://docs.nvidia.com/cuda/}{https://docs.nvidia.com/cuda/}}
\footnotetext[2]{\href{https://github.com/cgtuebingen/ggnn}{https://github.com/cgtuebingen/ggnn}}
\footnotetext[3]{\href{https://nvlabs.github.io/cub/}{https://nvlabs.github.io/cub/}}

\textbf{Push:}
For a given pair of index $\P$ and distance $\D$, the \Push method performs a parallel insertion into the distance-sorted lists of \Best and \PrioQ. %
To perform the insertion, all items with index $c_i$ that have a distance which is greater than $\D$ are temporarily copied into the thread-block's registers
and are written back into the subsequent index $c_{i+1}$ unless the end of the list or ring-buffer is reached. %
The new element is inserted where the item to the left is closer than the new item, or at the beginning if the new item is closer than all previous items. %
Consequently, a point that is eligible for $\Best$ and $\PrioQ$ will be inserted in both lists simultaneously. %

Since \PrioQ is a ring-buffer, the logical beginning and end positions are dependent on the head position. %
Hence, index computations on the physical borders need to be wrapped around. %

\textbf{Pop:}
This single-threaded routine returns the current head of the priority queue and manages the ring-buffers. %
Our \Criteria is monotonically decreasing over the course of a query (inspect Section~\ref{subsec:stopping}). %
Hence, if the head of the \PrioQ violates the \Criteria, we are able to safely terminate the query. %
In the valid case, the point is removed from the \PrioQ and added to the \Visited list at the head position. %
Both head pointers are moved one step forward. %
Finally, the point is returned. %

\subsubsection{Stopping Criterion}
\label{subsec:stopping}
On high-dimensional data, algorithms that solely rely on greedy downhill search will quickly get stuck in local minima. %
On the other hand, complete backtracking might visit the entire dataset.

A common stopping criterion, e.g. \cite{malkov2018efficient}, is to terminate the search once the \Best list cannot be improved by adding the closest not yet visited element,
i.e., once $d > d_{\text{best}_K}$, where $d$ is the distance of the new element and $ d_{\text{best}_K}$ is the distance of the last element in the \Best list.
To achieve higher recalls, the \Best list needs to be extended ($K$ needs to be increased), potentially by hundreds of elements.

\begin{figure}[htb]
    \centering
    \includegraphics[width=0.40\linewidth]{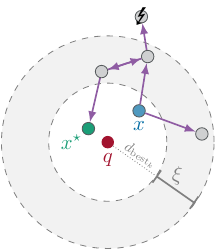}
    \hfill
    \includegraphics[width=0.53\linewidth]{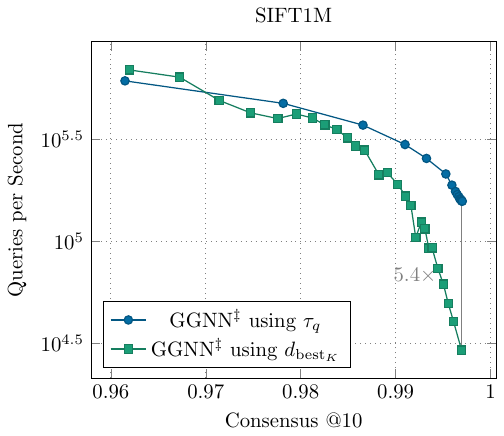}
    \caption{Allowing a slack $\xi$ makes it possible to escape a local minimum $x\in\mathcal{X}$, eventually reaching the solution $x^\star\in\mathcal{X}$ for a given query $q\in\mathbb{R}^d$.
    Points outside of this boundary are unlikely to provide helpful links and are discarded to reduce computational cost.
    (right) Comparison of using our stopping criterion ($\tau_q$) to using a growing \protect\Best list to terminate queries.
    With the \protect\Best list based approach, high recalls can only be achieved in conjunction with high memory usage which slows down the query.}
    \label{fig:slack}
\end{figure}

We propose an efficient approximation by adding an adaptive, monotonically decreasing slack $\xi$ to the distance of the $k$-closest neighbor: $d_{\text{best}_K} \approx d_{\text{best}_k} + \xi$, where $k \ll K$, that allows us to limit the size of the \Best list to only the $k$ closest neighbors which are to be queried (or at least 10).
Instead of explicitly tracking the distance of further neighbors, the slack accounts for a safety margin to ensure that the path to the closest neighbor will likely be found, e.g.\ Figure~\ref{fig:slack}.
The search will terminate once
\begin{equation}
\label{eq:stopping}
d > d_{\text{best}_k} + \xi,
\quad
\xi = \tau \cdot \min\{d_{\text{best}_1}, d^+_{\text{nn}_{1}}\}.
\end{equation}
The slack factor $\tau$ controls the size of the safety margin. %
$d_{\text{best}_1}$ is query-specific and relates the margin to the currently best found match %
while $d^+_{\text{nn}_{1}}$ correlates with the density of the given database nodes.
Specifically, $d^+_{\text{nn}_{1}}$ provides a global limit (to catch outliers) that is calculated during graph construction.
It denotes the maximum distance to the closest neighbor across all points within the currently processed subset of the graph $\mathcal{S}$:
\begin{equation}
    d^+_{\text{nn}_{1}} = \max_{x \in \mathcal{S}} \left\{\min_{y\in \mathcal{N}_x} \|x-y\|_2 \right\}, \mathrm{where}~ \mathcal{S} \subseteq \mathcal{X}.
\end{equation}

Using our stopping criterion keeps the \Best list at a small size and by that reduces the amount of shared memory required for performing the query.
As demonstrated by Figure~\ref{fig:slack}, our stopping criterion allows for high efficiency even at high accuracy.

Typical values for $\tau$ are above 0 and below 2.
Larger values improve accuracy with diminishing returns.

\section{Approximate Symmetric Nearest Neighbor Graph Construction}
\label{sec:method}
The construction of proper CPU-based kNN search-graphs is performed typically in a sequential procedure on global graphs. %
Edges are either attached (e.g., \cite{malkov2018efficient}), or pruned (e.g., \cite{harwood2016fanng,li2019approximate,fu2019fast}) one after the other. %
During the process, the edge-lists per node tend to vary heavily ranging from completely empty to complete lists of all possible points. %
While these global optimization approaches are able to produce high quality graph structures on CPUs, for fast GPU-based construction, they are impractical. %

We propose a parallel graph construction based on parallel merging of hierarchical kNN search-graphs as sketched in Figure~\ref{fig:merging_new}.
By partitioning the task of constructing the search graph into small, parallelizable tasks, we are able to efficiently use the massive parallism offered by GPUs.
In the following, we present our parallel graph construction process, followed by details on the hierarchical query process, and the graph-diversification and refinement by symmetric linking.
Finally, we look at how to best perform the final queries and multi-GPU configurations.

\subsection{Building the Hierarchical kNN-Graph}
\label{subsec:construction}
Our construction process (Algorithm~\ref{alg:parallel}) builds the kNN search-graph bottom-up by recursively merging smaller search-graphs.

To {\Init}ialize the process, we logically partition the entire dataset $\mathcal{X}$ into small batches of size $s$, e.g. 32.
These represent hierarchical search-graphs of height 1.
We initialize the neighbors within the bottom layer by first performing the \Merge operation which initializes the $k$ outgoing edges per point with each point's nearest neighbors.
Then, the \Sym operation replaces up to $k_{sym}$ outgoing links in an effort to approximate an undirected graph, details in Section~\ref{subsec:symmetry}.

In order to merge the individual sub-graphs, we first \Select $s$ points from the groups of $g$ graphs which each are to be merged.
These selected points now form batches representing the new top layer. %
There, we perform the same operations as with the bottom-layer before.
We {\Merge}, i.e. query for the nearest neighbors to fill the outgoing edge list and perform graph-diversification (\Sym).
As the nearest neighbor of any point might be found in any sub-graph, edges across batch boundaries will be introduced.
These steps now repeat top-down until the entire hierarchical search-graph is interconnected.

\begin{algorithm}[tb]
\caption{Parallel graph construction.}
\label{alg:parallel}
\Fn{\BT{\Base, \L, \K, \XIb}}{
    \KwIn{Dataset \Base, number of layers \L, number of neighbors per point \K, slack factor \XIb}
    \KwResult{search structure \Tree}
    \Tree $\leftarrow$ \Init{\Base, \L, \K, \XIb}\; %
    \For{\LT $\leftarrow 0$ \KwTo $\L-1$}{
        \If{\LT$> 0$}{
            \Tree.\Select{\LT}\;
        }
        \For{\LB $\leftarrow$ \LT \KwTo ~$0$}{
            \Tree.\Merge{\LT, \LB}\;
            \Tree.\Sym{\LB}\;
        }
    }
}
\BlankLine
\Fn{\Tree.\Merge{\LT, \LB}}{
    \ParFor{$\N \in \Tree.\Tlayer{\LB}$}{
        \tcc*[h]{Use brute force if $\LT = \LB$.}\;
        $\Buffer_{\N} \leftarrow$ \Tree.\Query{\N, \LT, \LB}\;
    }
    $\Tree.\Tlayer{\LB} \leftarrow \Buffer$\;
    \If{$\LB = 0$}{
        \Tree.\Stats{}\;
    }
}
\BlankLine
\Fn{\Tree.\Sym{\Li}}{
    \ParFor{$\N \in \Tree.\Tlayer{\Li}$}{
        \Tree.\SymQuery{\N}\;
    }
}
\end{algorithm}

Whenever the bottom layer is reached, we also save the distance to the first nearest neighbor $d_{\text{nn}_{1}}$ for each point and compute the mean and max over the dataset (\Stats). %
This allows for an easy adaption of our stopping criterion to every dataset without any precomputation. %
Initially, the maximum is prone to outliers in the still coarse nearest neighbor graph.
Therefore, during construction, we substitute the maximum distance to the closest neighbor $d^{+}_{\text{nn}_{1}}$ in Eq.~\ref{eq:stopping} with the mean distance~$\bar{d}_{\text{nn}_{1}}$:
\begin{equation}
    \bar{d}_{\text{nn}_{1}} = \frac{1}{|\mathcal{S}|} \sum_{x \in \mathcal{S}} \left\{\min_{y\in \mathcal{N}_x} \|x-y\|_2 \right\}, \mathrm{where}~ \mathcal{S} \subseteq \mathcal{X}.
\end{equation}

For selecting the points for the top layer, weighted reservoir sampling with $d_{\text{nn}_{1}}$ as weights is used with the method of \citet{EFRAIMIDIS2006181}.

As all construction tasks can be performed in parallel for each point in each layer while requiring small, limited amounts of memory each,
the construction can be performed quite quickly when exploiting the massive parallelism offered by GPUs.

\paragraph*{Hierarchical kNN Graph Query}
\label{subsec:hierarchical_query}
In a top layer, the \Merge operation can simply perform brute-force search between the $s$ points per batch.
With each lower layer, the number of batches grows with the factor $g$.
To perform efficient searches for nearest neighbors in lower layers we perform a hierarchical query (Algorithm \ref{alg:merge}),
which layer by layer utilizes the already merged upper layer to find entry points into the batches on the lower layers, and otherwise behaves just like the previously presented simple query (Section~\ref{subsec:query}).
The indices of the closest points found on the upper layer are then {\Transform}ed into the indices of the same points on the next lower layer
and the search continues until the target layer $l_m$ is reached.
While distances to already visited points can be reused, their neighborhood on the lower layers changes.
Therefore, we allow all points to be visited again after switching between layers.

As in HNSW~\cite{malkov2018efficient}, the hierarchy allows us to bridge gaps in the connectivity as multiple entry points from the top layer may approach the query on the bottom layer from different sides.

\begin{algorithm}[tb]
\caption{Hierarchical query.}
\label{alg:merge}
\Fn{\Tree.\Query{\Z, \LT, \LB}}{
    \KwIn{query point \Z from layer \LB, start layer \LT, and end layer \LB}
    \KwOut{\K closest points to \Z}
    \Cache.\Init{\Z, \XIg}\;
    \Sbf $\leftarrow$ \TopSeg{\Z, \LT}\;
    \Cache.\Fetch{\Sbf}\;
    \For{$ \Li \leftarrow l_{top}-1$ \KwTo \LB}{
        \Cache.\Transform{\Li}\;
        \While{$(\A \leftarrow \Cache.\Pop{}) \neq \emptyset$}{
            \Cache.\Fetch{$\mathcal{N}_{\A}$}\;
        }
    }
    \Return \Cache.\Best\;
}{}
\end{algorithm}

\subsection{Graph Diversification}
\label{subsec:symmetry}
In order to reach any point from any direction, every edge of the graph, in principle, should be undirected.
However, when every point should at least know its $k_\text{nn}$ nearest neighbors,
the total number of edges per point $x$ would vary significantly for an undirected graph as the number of points which have $x$ as their nearest neighbor will heavily depend on the local geometry. %

In order to obtain a regular representation and to allow for simple parallelized traversal with constant workload per step, our graph only consists of exactly $k$ directed (\ie, outgoing) edges.
We logically split the outgoing edges into at least $k_\mathrm{nn} = k/2$ true nearest neighbors and up to $k_\mathrm{sym} = k/2$ inverse links that are used to approximate an undirected graph.

As the number of representable inverse links is limited, one has to determine which of the inverse links are necessary.
Harwood and Drummond~\cite{harwood2016fanng} introduce the concept of shadowed links which are made redundant by the query's greedy exploration strategy. %
Optimizing the entire graph to remove all potentially shadowed edges requires a complex global optimization. The same holds for the construction of monotonic relative neighborhood graphs~\cite{fu2019fast}. %

\begin{figure}[htb]
    \centering
    ~\hfill
    \includegraphics[height=0.3\linewidth]{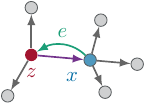}
    \hfill
    \includegraphics[height=0.3\linewidth]{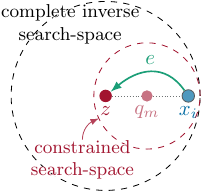}
    \hfill~
    \caption{Maintaining symmetric links. If there is no easily found connection from $x\in\mathcal{N}_z^{nn}$ back to $z\in\mathcal{X}$, the edge $e$ is added to allow propagating nearest neighbor information between $z$ and $x$.
    (right) Instead of considering the full search radius when probing whether to insert symmetric links, we constrain the maximum distance of points to be visited to the smaller sphere centered around the midpoint $q_m$.}
    \label{fig:symmetric_links}
\end{figure}

Our approach to graph diversification is to explicitly search for missing inverse links by querying for each point $z$ from all its $k_\mathrm{nn}$ nearest neighbors $x_i \in\mathcal{N}_z^{nn}$ within a small search radius.
Where a direct inverse link from $x_i$ back to $z$ exists, the query can trivially be skipped.
When there is no path within the allowed range, the link is added (e.g., Figure~\ref{fig:symmetric_links}).
The searches are executed in parallel over all points $z \in \mathcal{X}$, allowing for faster construction.

\begin{algorithm}[ht]
\caption{Symmetric linking query.}
\label{alg:sym_links}
\Fn{\Tree.\SymQuery{\Z}}{
\KwIn{symmetric query point \Z}
\ForEach{$x_i \in \mathcal{N}^{nn}_{\Z}$}{
\Cache.\Init{\Z, \XIg}\;
\Cache.\Fetch{$x_i$}\;
\While{$(\A \leftarrow \Cache.\Pop{}) \neq \emptyset$}{
    \Pbf $\leftarrow$ $\mathcal{N}^{nn}_{\A} \cup \mathcal{N}^{sym}_{\A}$\;
    \If{\Z $\in$ \Pbf}{
        skip $x_i$\;
    }
    \Cache.\Fetch{\Pbf}\;
}
\ForEach{$\P \in \Cache.\Best$}{
    \If{$ \left| \mathcal{N}^{sym}_{\P} \right| < \KF $}{
        $\mathcal{N}^{sym}_{\P} \leftarrow \mathcal{N}^{sym}_{\P} \cup \{\Z\} $\;
        \Break\;
    }
}
}
}
\end{algorithm}

The detailed symmetric linking operation is shown in Algorithm~\ref{alg:sym_links}.
Each point $z$ launches queries for itself, starting from its $k_\mathrm{nn}$ nearest neighbors $x_i$.
During the query, only the $k_\mathrm{nn}$ nearest neighbors of each visited point $\mathcal{N}_{x_i}^{nn}$
and potential already inserted symmetric/inverse links at that point $\mathcal{N}_{x_i}^{sym}$ are considered by the query as all others may still be overwritten.
If a path back to $z$ is found, the search continues with $z$'s next neighbor.
Otherwise, a link to $z$ is inserted at the closest point to $z$ encountered during the search which still has free capacities in its inverse link list.

To avoid race conditions on the insertions, we use atomics to keep track of the inverse link list sizes.
On average, less than $k/4$ inverse links are necessary this way.
If no candidate is found, the link is ignored.
In our setting, this is, however, a very rare case and $z$ might potentially still be reachable through either through hierarchy or multiple start points.

\subsubsection{Additional Symmetric Linking Constraint} \label{subsubsec:sym_link_constraint}
As demonstrated in Figure~\ref{fig:symmetric_links}, the usual search-space would be defined depending on $z$ and $x_i$. %
However, in order to generate a search structure that is able to find a path for all potential query points, we additionally constrain it on the worst-case query that still needs to find a path to $z$. %
Hence, a link is required to bridge the gap for a worst-case query on the midpoint $q_m$. %
Note, a path to $x_i$ is given since it is directly known from $z$. %
For symmetric linking, only this search space is relevant, which is the circle defined by the center $q_m$ and distance to $x_i$, but still encloses $z$.
In practice, we set $q_m = z + 0.4 \cdot (x_i - z)$.
This is slightly less conservative, but decreases the number of necessary symmetric links.

\subsection{Graph Refinement} \label{subsec:graph_refinement}
The initial graph construction might not always produce perfect results at the first trial.
The reason for this is to some extent the large merge factor $g$ that results when constructing search-graphs of low height that need to span over massive datasets.

In our experiments we found using a height of $L=4$ for the hierarchical search-graph to provide the best performance.
To create a single graph from the initially $n/s$ disjoint sub-graphs consisting of single batches of size $s$ (usually, $s=32$),
we need to merge groups of $g=\sqrt[L-1]{n/s}$ sub-graphs on each layer, where each new sub-graph's top-layer contains $s$ points sampled from these $g$ sub-graphs (see Section~\ref{subsec:construction}).
While for $n = 256$, just 2 graphs need to be merged together each, for $n = 10^6$, about 32 graphs each need to be merged per layer.
The higher $g$, the fewer entry points into the lower level exist, limiting the success rate of the initial hierarchical queries, as there are only $s/g$ per merged sub-graph.
For $g > s$, some points can only be reached after sub-graphs are bridged by symmetric linking.

To improve the graph quality, refinement steps, as described in Algorithm~\ref{alg:refine}, can be performed
which simply repeat the merging and symmetric linking steps throughout the search-graph.
While refinement steps would also be beneficial during the actual sub-graph merging, we observed that performing them in the end is sufficient for the overall quality of the search structure. %

\begin{algorithm}[ht]
\caption{Parallel graph refinement.}
\label{alg:refine}
\Fn{\Refine{\Tree, \LT}}{
\KwIn{search structure \Tree, top layer \LT}
\KwResult{refined search structure \Tree}
\For{\LB $\leftarrow$ $\LT-1$\KwTo $0 \;$}{
    \Tree.\Merge{\LT, \LB}\;
    \Tree.\Sym{\LB}\;
}
}
\end{algorithm}

\subsection{Query Considerations}
\label{subsec:querystart}
For our approach, during graph construction, it is crucial to perform a hierarchical query, which searches nearest neighbors layer by layer, due to the fact that the search index is not yet fully merged. %
Here, the coarse-to-fine methodology is able to bridge gaps between sub-graphs that are not yet connected and provides multiple routes to the area of interest. %
Starting from multiple spread-out points increases the quality significantly. %
In particular, if not only the 1-NN is of importance but also the neighbors at the far end of $k$.
However, for a converged search-graph, this effort is no longer necessary as all points are interlinked well. %

Searching in flat index structures does not have the hierarchical overhead. %
For example, Li~\etal \cite{li2019approximate} start their query always from the centroid in a flat graph. %
Nevertheless, as the number of points grows, the number of hops to the centroid grows as well. %
In addition, with a single starting point, any goal point will be approached from just one direction. %
The search might be prone to gaps or linking problems inside the search-graph. %

In our design, the bottom layer is actually a flat graph that has all points included. %
When searching the finished search-graph for nearest neighbors of novel points, it turned out to be more efficient in practice to skip the intermediate layers
and continue the search directly on the bottom layer after exploring the $s$ start points that make up the top layer.
This allows to significantly cut back on the search iterations performed as otherwise the search would need to iterate on each intermediate layer.
In practice, compared with the hierarchical approach, we did not observe any loss in recall but a significant saving in time. %

\subsection{Multi-GPU}
\label{subsec:multi_gpu}
The parallel construction and search algorithm can be extended to multiple GPUs.
Johnson \etal \cite{johnson2019billion} distinguish between two types of multi-GPU parallelism: \emph{Replication} and \emph{Sharding}.
\textit{Replication} copies the entire dataset $\mathcal{X}$ to multiple GPUs to parallelize the query process even further.
The queries $\mathcal{Q}$ are divided equally among the available GPUs, resulting in linear speedups.

\textit{Sharding} subdivides the dataset $\mathcal{X}$ into smaller, independent shards.
This allows us to process billion-scale datasets which would otherwise not fit into GPU memory.
When necessary, shards can also be swapped into main memory or onto disk.
All GPUs construct the search-graphs for their shards in parallel.
Since each shard has lower complexity due to the reduced number of points, super-linear speedups \wrt the same quality can be achieved.

The downside is that all shards need to be processed for each query.
When querying multiple shards per GPU, we swap already processed shards for new ones in the background to hide memory latencies.
Still, swapping shards puts a limit on the minimum run time per query batch.
As some additional overhead,
the individual query results per shard need to be merged.
Within each GPU, we use a parallel radix sort.
The final result across the GPUs is computed on the CPU using an efficient n-way merge.

%% file: tex/evaluation.tex
\section{Empirical Evaluation}
\label{sec:evalution}
In the following, we evaluate the performance of the proposed approach and its individual components on several publicly available benchmark datasets and report qualitative and quantitative results in terms of timings and accuracy.

\textbf{Datasets.}
We ran experiments on datasets of varying sizes and dimensionality generated in different application contexts:
SIFT1M~\cite{PQ} and SIFT1B~\cite{1BPQ} containing SIFT vectors of dimension 128, DEEP1B~\cite{Babenko2016EfficientIO} with
 1 billion 96-dimensional feature vectors encoding entire images, NyTimes~\cite{aumueller2018annbenchmarks,Dua:2019} and GloVe 200~\cite{aumueller2018annbenchmarks,pennington2014glove}, which are two skewed and clustered datasets containig random projections of word embeddings compared using cosine-similarity, and finally the 960-dimensional dataset GIST~\cite{PQ}.
Details on the datasets and construction details for our search graphs can be found in Table~\ref{tab:construction_times}.

\textbf{Hardware.}
We use a machine with 8x NVIDIA Tesla V100 and 2x Intel Xeon Gold 5218,
where we usually use just a single GPU, except for SIFT1B and DEEP1B, where we use all 8 GPUs.
We publish results for additional GPUs in the supplemental.

\textbf{Performance Metrics.}
When comparing results to other approaches, one has to carefully look at the employed metric.
The query performance is typically measured in recall (R@k).
For quantization-based methods (e.g.,~\cite{PQ}), the recall is understood as the ratio of queries which return the true first nearest neighbor among the first $k$ results:
\begin{align}
    R@k=\frac{\left\vert \mathcal{N}_q^\text{gt}(1) \cap \mathcal{N}_q(k)\right\vert}{\left\vert \mathcal{N}_q^\text{gt}(1) \right\vert}.
\end{align}

For methods with exact distance computations -- such as graph-based methods -- the recall is instead understood as the overlap between the first $k$ results and the first $k$ true nearest neighbors.
For disambiguation, we refer to it as consensus (C@k):
\begin{align}
    C@k=\frac{\left\vert \mathcal{N}_q^\text{gt}(k) \cap \mathcal{N}_q(k)\right\vert}{\left\vert \mathcal{N}_q^\text{gt}(k) \right\vert}.
\end{align}

As our algorithm uses exact distance calculations, the true nearest neighbor is either reported as the first element in an answer or not found at all.
Therefore, we report only R@1 ($=$C@1).

\textbf{Batch Size.}
The number of queries executed per batch is an important factor for GPU-based queries,
which we analyze in Figure~\ref{subfig:batchsize}.
In our experiments, we measure the time for executing all 10000 queries (except 1000 queries for GIST~\cite{PQ}) in a batch
and report the total execution time divided by the number of queries as $\si{\micro\second}$/query
or plot it as queries per second. %

\begin{table}[htbp]
  \caption{Million-scale performance comparison.}
  \label{tab:million_comparision}
  \centering
  \begin{tabular}{lcS[detect-weight,table-format=5.1]ccc}
  \toprule
    & & \multicolumn{4}{c}{SIFT1M~\cite{PQ}}
    \\
    \cmidrule{3-6}
    \multirow{2}{*}{Approach}
    & \multirow{2}{*}{$\tau_q$} & {Query time} & {Recall}  & {Recall} & {Recall}
    \\
    & & {\si{\micro\second}/query $\downarrow$} & {@1 $\uparrow$} & @10 $\uparrow$ & @100 $\uparrow$
    \\
    \midrule
    LOPQ~\cite{localPQ}
    & & 51100 & 0.51 & 0.93 & 0.97  %
    \\
    IVFPQ~\cite{PQ}
    & & 11200 & 0.28 & 0.70 & 0.93  %
    \\
    FLANN~\cite{flann_pami_2014}
    & & 5320 & 0.97 & {-} & {-}     %
    \\
    \multirow{2}{*}{HM-ANN~\cite{jieren2020hm-ann}} %
    & & 855 & 0.9995 & {-} & {-}
    \\
    & & 236 & 0.99 & {-} & {-}
    \\
    \midrule
    PQT~\cite{wieschollek2016efficient}
    & & 20 & 0.51 & 0.83 & 0.86
    \\
    FAISS~\cite{johnson2019billion}
    & & 20 & 0.80 & 0.88 & 0.95
    \\
    PQFPGA~\cite{PQFPGA}
    & & 20 & 0.88 & 0.94 & 0.97
    \\
    \multirow{2}{*}{FANNG~\cite{harwood2016fanng}} %
    & & 1.3 & 0.95 & {-} & {-}
    \\
    & & 0.8 & 0.90 & {-} & {-}
    \\
  \midrule
  \multirow{4}{*}{GGNN} %
  & 0.65 & 5.8 & \textbf{0.9997} & {-} & {-}
  \\
  & 0.42 & 1.5 & 0.99 & {-} & {-}
  \\
  & 0.30 & 0.7 & 0.95 & {-} & {-}
  \\
  & 0.20 & \bfseries 0.5 & 0.90 & {-} & {-}
  \\
  \midrule
  \multicolumn{2}{c}{Brute Force (1x V100)}
  & 381.4 & 1.0 & {-} & {-}
  \\
  \bottomrule
  \end{tabular}
\end{table}

\begin{table}[htbp]
  \caption{Performance comparison on SIFT1B~\cite{1BPQ}.
  Our method GGNN$^8$ used 8 GPUs. RobustiQ~\cite{chen2019robustiq} used 2 GPUs. Otherwise, single CPUs or GPUs were used.
  Our method achieves perfect recall@1.
  GGNN$^8$ is limited by the host to device memory bandwidth and could otherwise perform the query with 99\% recall@1 in just 24.5\si{\micro\second}/query (see Section \ref{subsec:out_of_memory}).
  The smaller GGNN$^{8\ddagger}$ can be queried much faster, since it iterates over fewer shards and fits on the GPUs in one piece.}
  \label{tab:sift1b_comparision}
  \centering
  \begin{tabular}{l@{}cS[detect-weight,table-format=5.1]ccc}
    \toprule
    & & \multicolumn{4}{c}{SIFT1B~\cite{1BPQ}}
    \\
    \cmidrule{3-6}
    \multirow{2}{*}{Approach}
    & \multirow{2}{*}{$\tau_q$} & {Query time} & {Recall}  & {Recall} & {Recall}
    \\
    & & {\si{\micro\second}/query $\downarrow$} & {@1 $\uparrow$} & @10 $\uparrow$ & @100 $\uparrow$
    \\
    \midrule
    LOPQ~\cite{localPQ}
     & & 8000 & {-} & 0.20 & {-}
     \\
     IVFPQ~\cite{PQ}
     & & 74000 & 0.08 & 0.37 & 0.73
     \\
     Multi-D-ADC~\cite{invertedmultiindex}
     & & 1603 & 0.33 & 0.80 & 0.98 %
     \\
     HM-ANN~\cite{jieren2020hm-ann} %
     & & 1014 & 0.99 & {-} & {-}
    \\
    \midrule
     PQT~\cite{wieschollek2016efficient}
     & & 150 & 0.14 & 0.35 & 0.57
     \\
     FAISS~\cite{johnson2019billion}
     & & 17.7 & {-} & 0.37 & {-}
     \\
     PQFPGA~\cite{PQFPGA}
     & & 20 & {-} & 0.55 & {-}
     \\
     RobustiQ~\cite{chen2019robustiq}
     & & 33 & 0.33 & 0.76 & 0.90
     \\
    \midrule
    \multirow{3}{*}{GGNN$^8$}
    & 0.56 %
    & 61.6 & \bfseries 1.0 & {-} & {-}
    \\
    & 0.50 %
    & 46.4 & 0.999 & {-} & {-}
    \\
    & 0.38 %
    & 42.7 & 0.99 & {-} & {-}
    \\
    \midrule
    \multirow{2}{*}{GGNN$^{8\ddagger}$}
    & 0.61 %
    & 7.1 & 0.99 & {-} & {-}
    \\
    & 0.42 %
    & \bfseries 3.9 & 0.90 & {-} & {-}
    \\
    \midrule
    \multicolumn{2}{c}{Brute Force (8x V100)}
    & \!\!\SI{\approx 30000}{} & 1.0 & {-} & {-}
    \\
  \bottomrule
  \end{tabular}
\end{table}

\begin{table}[htbp]
  \caption{Performance comparison on DEEP1B~\cite{Babenko2016EfficientIO}.
  Our method GGNN$^8$ used 8 GPUs. FAISS~\cite{johnson2019billion} used 4 GPUs.
  RobustiQ~\cite{chen2019robustiq} used 2 GPUs.
  Otherwise, single CPUs were used.
  GGNN$^8$ is limited by the host to device memory bandwidth to at least 233\si{\micro\second}/query on the 10k queries set (see Section \ref{subsec:out_of_memory}).
  To show the influence of $\tau_q$ and to compare with future hardware configurations, we report times without the currently necessary upload as GGNN$^{8\star}$.}
  \label{tab:deep1b_comparision}
  \centering
  \begin{tabular}{l@{}cS[detect-weight,table-format=5.1]ccc}
    \toprule
    & & \multicolumn{4}{c}{DEEP1B~\cite{Babenko2016EfficientIO}}
    \\
    \cmidrule{3-6}
    \multirow{2}{*}{Approach}
    & \multirow{2}{*}{$\tau_q$} & {Query time} & {Recall}  & {Recall} & {Recall}
    \\
    & & {\si{\micro\second}/query $\downarrow$} & {@1 $\uparrow$} & @10 $\uparrow$ & @100 $\uparrow$
    \\
    \midrule
     Multi-D-ADC~\cite{invertedmultiindex}
     & & 1065 & 0.37 & 0.74 & 0.92 %
     \\
     HM-ANN~\cite{jieren2020hm-ann} %
     & & 1142 & \bfseries 0.99 & {-} & {-}
    \\
    \midrule
     FAISS~\cite{johnson2019billion}
     & & 13.3 & 0.45 & {-} & {-}
     \\
     RobustiQ~\cite{chen2019robustiq}
     & & 30 & 0.38 & 0.75 & 0.89
     \\
    \midrule
    \multirow{3}{*}{GGNN$^{8\star}$}
    & 1.50 %
    & 27.6 & \bfseries 0.99 & {-} & {-} %
    \\
    & 0.44 %
    & 8.3 & 0.95 & {-} & {-} %
    \\
    & 0.36 %
    & \bfseries 7.2 & 0.90 & {-} & {-} %
    \\
  \bottomrule
  \end{tabular}
\end{table}

\begin{table*}[p]
  \caption{Datasets and search graph construction details.
  For some datasets, we also demonstrate the construction-query trade-off with a faster construction version ($\ddagger$).
  All configurations shown here are capable of reaching at least 99\% recall@1.}
  \label{tab:construction_times}
  \centering

  \begin{tabular}{lrr @{\,}c@{~~}c@{~~} lccccc @{\,}c@{~~}c@{~~} ccr@{~}lr}
  \toprule
  \multicolumn{3}{c}{Dataset} & & & \multicolumn{6}{c}{Search Graph Configuration} & & & \multicolumn{5}{c}{Construction Time and Graph Size} \\
  \cmidrule{1-3} \cmidrule{6-11} \cmidrule{14-18}
  Dataset & \multicolumn{1}{c}{\#Points} & \#Dim. & & & Config. & $k$ & $s$ & $l$ & $\tau_b$ & $r$ & & & \multicolumn{2}{c}{GPU} & \multicolumn{2}{c}{Time} & \multicolumn{1}{c}{Size} \\
  \midrule
  \multirow{2}{*}{SIFT1M~\cite{PQ}}
  &       \multirow{2}{*}{$1\, 000\, 000$} & \multirow{2}{*}{128} & &
  & GGNN$^\ddagger$ & 24 & 32 & 4 & 0.5 & 2 & & & \multirow{6}{*}{1x V100} &           &  9.2 & s &  95 MiB
  \\
  &                                        &                      & &
  & GGNN            & 40 & 32 & 4 & 0.5 & 2 & & &                          &           & 21.8 & s & 158 MiB
  \\
  \multirow{2}{*}{NyTimes~\cite{aumueller2018annbenchmarks,Dua:2019}}
  &           \multirow{2}{*}{$290\, 000$} & \multirow{2}{*}{256} & &
  & GGNN$^\ddagger$ & 40 & 32 & 4 & 0.3 & 0 & & &                          &           & 13.4 & s &  46 MiB
  \\
  &                                        &                      & &
  & GGNN            & 96 & 64 & 4 & 0.3 & 0 & & &                          &           & 39.3 & s & 113 MiB
  \\
  GloVe 200~\cite{aumueller2018annbenchmarks,pennington2014glove}
  &                       $1\, 183\, 514$  &                 200  & &
  & GGNN            & 96 & 64 & 4 & 0.5 & 2 & & &                          &           &  6.0 & min & 450 MiB
  \\
  GIST~\cite{PQ}
  &                       $1\, 000\, 000$  &                 960  & &
  & GGNN            & 96 & 64 & 4 & 0.4 & 2 & & &                          &           &  11.9 & min & 382 MiB
  \\
  \midrule
  \multirow{2}{*}{SIFT1B~\cite{1BPQ}}
  & \multirow{2}{*}{$1\, 000\, 000\, 000$} & \multirow{2}{*}{128} & &
  & GGNN$^{8\ddagger}$ & 20 & 32 & 4 & 0.5 & 2 & & & \multirow{3}{*}{8x V100} &  16 shards & 33.4 & min &  75 GiB
  \\
  &                                        &                      & &
  & GGNN$^8$           & 40 & 32 & 4 & 0.5 & 2 & & &                          & 128 shards & 78.6 & min & 152 GiB
  \\
  DEEP1B~\cite{Babenko2016EfficientIO}
  &                 $1\, 000\, 000\, 000$  &                 96   & &
  & GGNN$^8$ & 24 & 32 & 4 & 0.5 & 2 & & &                                    &  32 shards & 47.5 & min &  90 GiB
  \\
  \bottomrule
  \end{tabular}
\end{table*}

\begin{figure*}[p]
    \centering
    \includegraphics[width=0.32\linewidth]{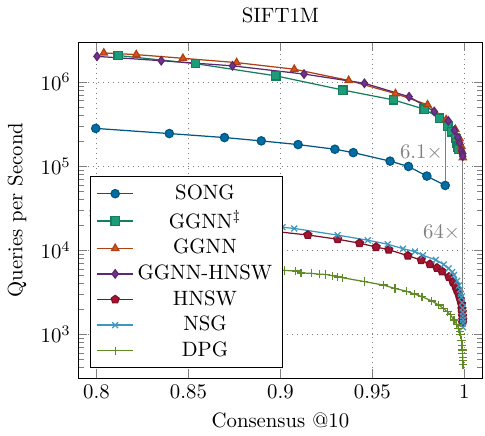}
    \hfill
    \includegraphics[width=0.32\linewidth]{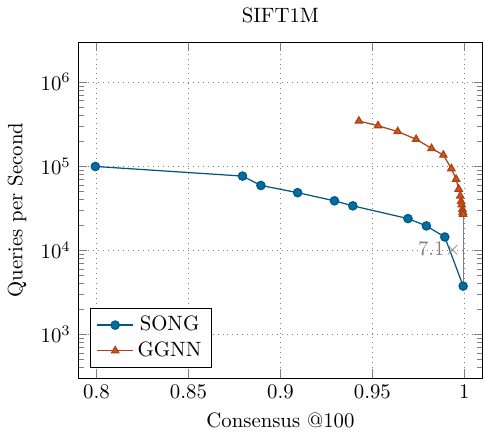}
    \hfill
    \includegraphics[width=0.32\linewidth]{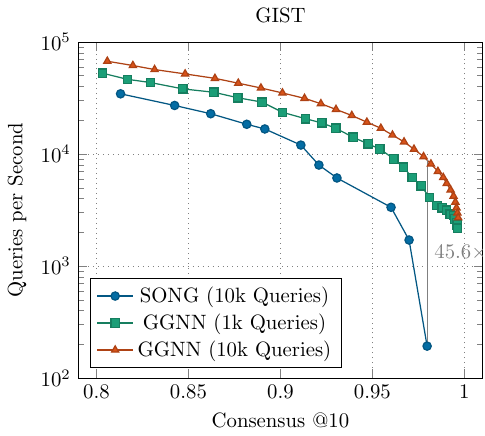}
    \\
    \includegraphics[width=0.32\linewidth]{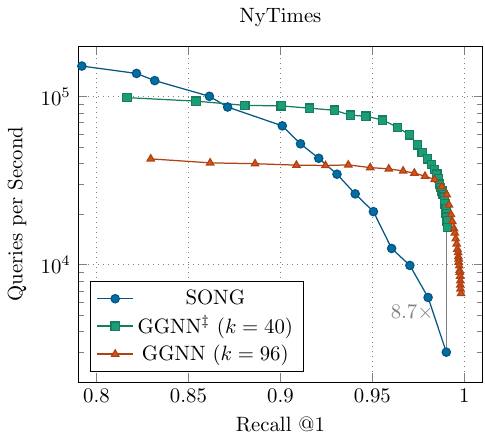}
    \hfill
    \includegraphics[width=0.32\linewidth]{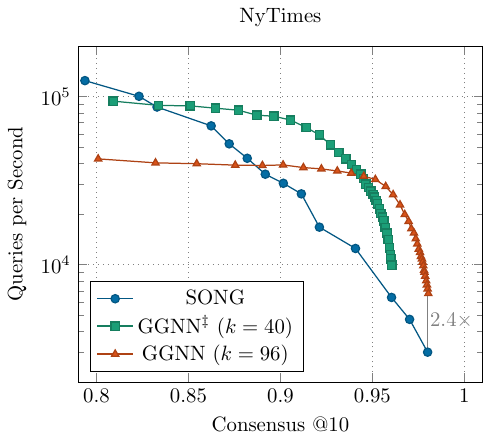}
    \hfill
    \includegraphics[width=0.32\linewidth]{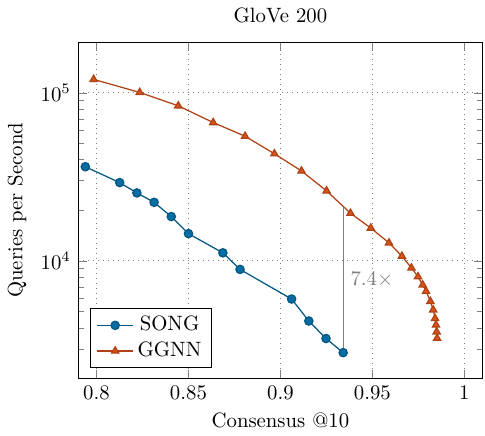}
    \\
    \includegraphics[width=0.32\linewidth]{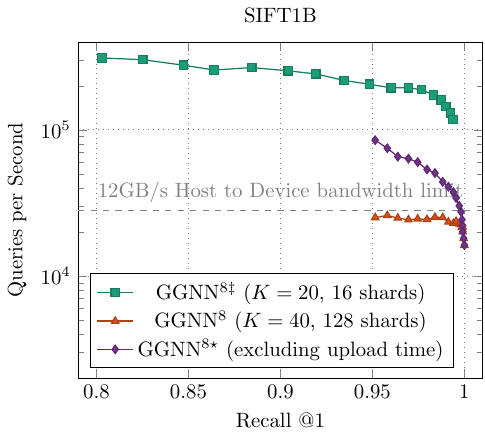}
    \hfill
    \includegraphics[width=0.32\linewidth]{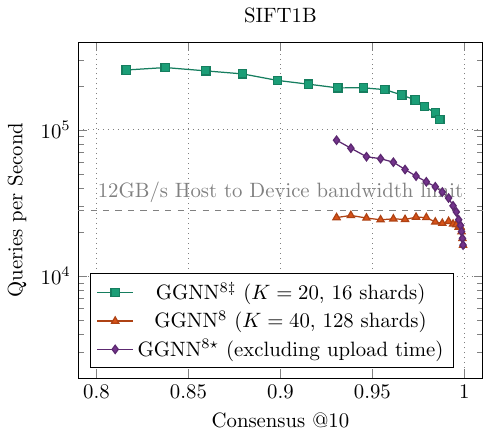}
    \hfill
    \includegraphics[width=0.32\linewidth]{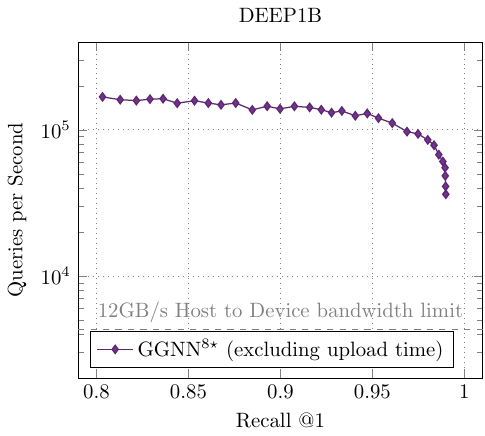}
    \caption{Performance comparison across various datasets and top-$k$ queries.
    Results in the top-right corner are preferable.
    We compare mainly against SONG~\cite{zhao2020song}, using the values from their paper and using the same GPU.
    On SIFT1M, we additionally compare against state-of-the-art single-core CPU-based methods~\cite{malkov2018efficient,li2019approximate,fu2019fast}
    and include a configuration "GGNN-HNSW", where we use our method to query the graph built by HNSW~\cite{malkov2018efficient}, like SONG~\cite{zhao2020song} does.
    Our query significantly outperforms SONG and achieves similar results as a query on a similarly-sized search-graph constructed in much less time using our method.
    High performance is also achieved when querying for the 100 nearest neighbors or more complicated datasets like GloVe 200 or very high dimensional datasets like GIST.
    On the rather small NyTimes dataset, our graph construction appears to be non-optimal but the highly efficient query still manages to outperform SONG by a large factor.
    Using 8 GPUs, even billion-scale datasets like SIFT1B or DEEP1B can be queried at rates around 100k queries per second with near-perfect recalls.}
    \label{fig:performance_comparisons}
\end{figure*}

\subsection{Performance Comparisons}
Evaluation is performed on the default configuration GGNN,
a faster, less accurate configuration GGNN$^\ddagger$,
and a 8-GPU configuration GGNN$^8$ (see Table~\ref{tab:construction_times} for parameters).

As reference, we use several recent CPU-based methods \cite{malkov2018efficient,li2019approximate,fu2019fast}, where we executed the single-threaded reference implementations on an Intel Core i7-9700K.
We further compare against the GPU-based SONG~\cite{zhao2020song} and mimic their approach of querying the bottom layer of a prebuilt HNSW~\cite{malkov2018efficient} graph using our query as "GGNN-HNSW".
Using our query is even faster. In addition, it performs very similar to querying a similarly sized graph constructed using our method in significantly less time.
Detailed performance comparisons are shown in Figure~\ref{fig:performance_comparisons} and Tables~\ref{tab:million_comparision}, \ref{tab:sift1b_comparision}, and \ref{tab:deep1b_comparision}.
In the tables, we compare against previously published results,
where GPU-based methods~\cite{wieschollek2016efficient,harwood2016fanng,johnson2019billion,chen2019robustiq} used NVIDIA GTX Titan X GPUs,
and \cite{PQFPGA} used an Arria 10 GX1150 FPGA.
Besides being significantly faster on all datasets (around $1\si{\micro\second}$ per query on SIFT1M), our method can also achieve very high recall rates, close to perfect.

\subsection{Search-Graph Construction}
In the following, we inspect several aspects of the construction process, including
the time spent on the individual construction steps, %
the influence of the dataset size, %
the resulting graph quality in terms of nearest neighbors per point, %
and how additional construction time can be traded in for even faster queries. %

\subsubsection{Construction Time}
Construction times and index sizes of our method are listed in Table~\ref{tab:construction_times}.
For comparison, FAISS~\cite{johnson2019billion} report construction times between 4 and 24 hours for DEEP1B, yet reaches less than 50\% R@1.
HM-ANN~\cite{jieren2020hm-ann} report construction times around 100 hours for effective billion-scale search-graphs.
Using our hierarchical GPU-based graph-merge algorithm, 99\% R@1 on DEEP1B and even perfect 100\% R@1 SIFT1B can be reached with fast queries
while the construction takes less than 2 hours.

\subsubsection{Construction Time Composition}
\label{subsec:construction_time_composition}
\begin{figure}[htb]
    \centering
    \includegraphics[width=0.95\columnwidth, trim={0 2pt 0 0}, clip]{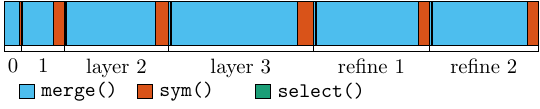}
    \caption{Split-up of the construction time for the SIFT1M search graph.
    Most time is spent on the \texttt{merge} operation (86.92\%), some time on the \texttt{sym} operation (13.06\%) and almost no time on \texttt{select}ing the points for the upper layers (0.02\%).}
    \label{fig:sift1m_construction}
\end{figure}

Figure~\ref{fig:sift1m_construction} shows the time spent on the individual graph-construction operations with 2 refinement iterations.
The majority of the construction time is spent on the \texttt{merge} operation, especially those merges which involve the bottom layer,
where the merge operation searches for the $k$ nearest neighbors for all points in the dataset.

\subsubsection{Dataset Size}
\label{subsec:dataset_size}
\begin{figure}[htb]
    \centering
    \includegraphics[height=0.45\linewidth]{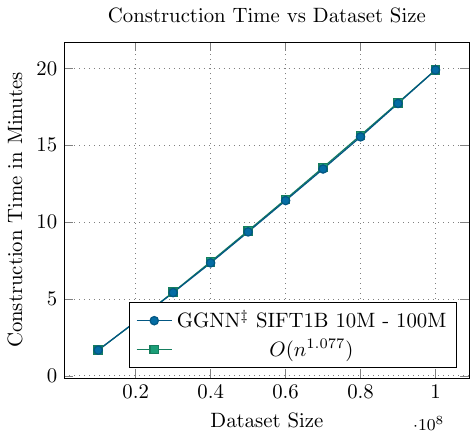}
    \hfill
    \includegraphics[height=0.45\linewidth]{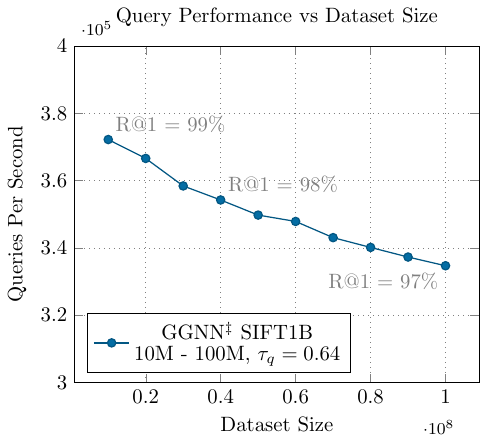}
    \caption{Construction times and queries per second for increasing subsets of SIFT1B (up to 100M) on a single GPU.
    The construction time almost linearly depends on the dataset size.
    With increasing dataset size, the query performance suffers slightly in both execution time and accuracy for a constant slack factor $\tau_q$.}
    \label{fig:sift1b_subsets_construction}
\end{figure}

Efficient graph construction needs to scale well with the size of the dataset.
As demonstrated in Figure~\ref{fig:sift1b_subsets_construction},
the construction time scales almost linearly with the size of the dataset, $\approx O(n^{1.077})$.
This can be explained by most of the construction time being spent on independently determining the neighbors of each bottom-level point (see Section~\ref{subsec:construction_time_composition}).
This almost linear behavior is significantly better than the complexity of brute-force kNN construction~$O(n^2)$.

\subsubsection{For-all Queries}
\label{subsec:for_all_queries}
The fast construction of the kNN search graph from scratch solves another interesting problem on the side,
namely computing the k nearest neighbors for all points in the dataset, as frequently encountered in n-body problems. %
When we construct the full hierarchy, the final merging step effectively searches for the k neighbors of all points. %
In Table~\ref{tab:forAll}, we demonstrate the quality of the for-all problem considering the consensus. %
Referring to Figure~\ref{fig:sift1b_subsets_construction}, this for-all problem again shows almost a linear behavior. %
In case of sharding, though, all points would need to be tested once against every shard. %
As the number of shards linearly depends on the number of points one approaches $O(n^2)$ complexity again with a tiny constant though, e.g.\ 16 shards for the SIFT1B dataset.

\begin{table}[th]
  \caption{Accuracy of the nearest neighbors for each point in the dataset, as contained within our search graph.}
  \centering
  \begin{tabular}{llcc}
    \toprule
    Method & Dataset & C@1 $\uparrow$ & C@10 $\uparrow$ \\
    \midrule
    GGNN$^\ddagger$ & \multirow{2}{*}{SIFT1M} & 0.996 & 0.981 \\
    GGNN & & 0.999 & 0.998 \\
    \bottomrule
  \end{tabular}
  \label{tab:forAll}
\end{table}

\subsubsection{The Construction-Query Trade-off}
\label{subsec:construction_query_trade_off}
\begin{figure}[t]
    \centering
    \begin{subfigure}[t]{0.48\linewidth}
        \centering
        \includegraphics[height=0.9\linewidth]{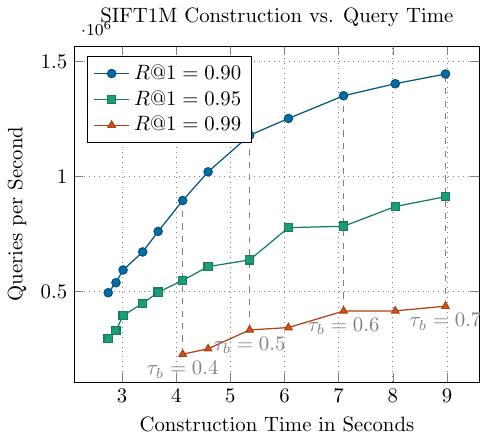}
        \caption{Varying construction time.}
        \label{subfig:build_vs_query}
    \end{subfigure}
    \hfill
    \begin{subfigure}[t]{0.48\linewidth}
        \centering
        \includegraphics[height=0.9\linewidth]{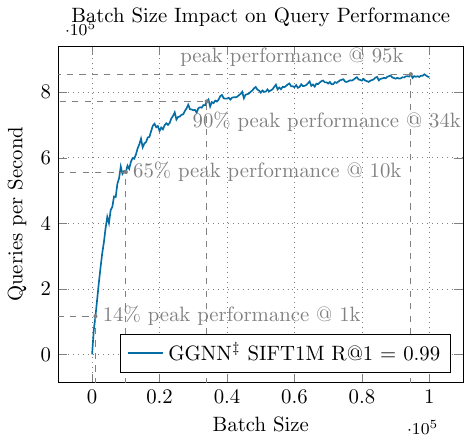}
        \caption{Varying batch size.}
        \label{subfig:batchsize}
    \end{subfigure}
    \caption{Query performance depends both on the quality of the search-graph and the number of queries executed simultaneously.
    Queries with the same accuracy can be executed faster if more time is spent on construction (\subref{subfig:build_vs_query}).
    Here, no refinement is performed ($r=0$) to visualize the influence of $\tau_b$ on the initial construction.
    Generally, having a few refinement iterations can further boost query performance.
    In terms of query size (\subref{subfig:batchsize}), high performance is already reached for a few thousand queries, while further simultaneous queries yield even better performance with a peak at 95k queries.}
    \label{fig:query_time}
\end{figure}

Our graph construction algorithm and the query can be tuned for speed or accuracy.
The build time is controlled by the slack factor $\tau_b$ and by how many refinement iterations $r$ are carried out. %
The longer the construction time, the more precise the resulting graph. %
A quickly assembled graph will have worse edges and a query might need to visit more nodes until fulfilling the stopping criterion. %
In Figure~\ref{subfig:build_vs_query}, the trade-off between build time and query time is visualized for fixed recall rates.

\subsection{Query Behaviour}
\label{subsec:query_behaviour}
The runtime and the quality of a query can be controlled by adapting the stopping criterion through the slack factor $\tau_q$. %
As can be seen e.g. in Table~\ref{tab:million_comparision}, by increasing $\tau_q$, a larger safety margin is considered during query,
resulting in better recall rates at the cost of visiting more points and consequently longer query time.

\begin{figure}[t]
   \centering
    \begin{subfigure}[t]{0.48\linewidth}
        \includegraphics[height=0.85\linewidth]{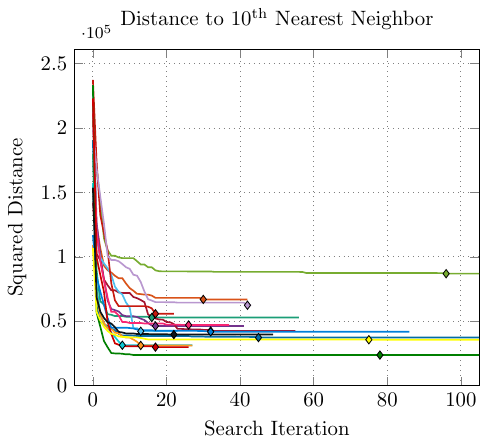}
        \caption{Development of the 10$^\mathrm{th}$ best query point as the query progresses.}
        \label{subfig:distance_per_iterations}
    \end{subfigure}
    \hfill
    \begin{subfigure}[t]{0.48\linewidth}
        \includegraphics[height=0.85\linewidth]{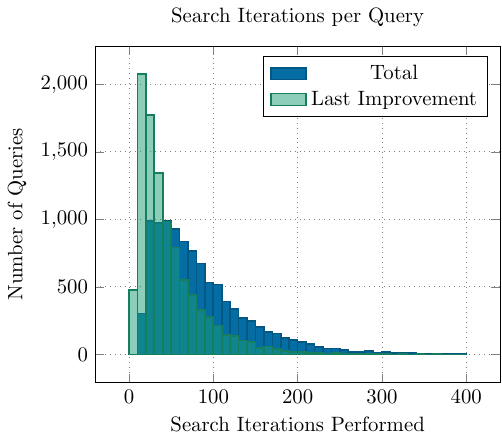}
        \caption{Iterations performed total, and until the last improvement is found.}
        \label{subfig:query_iterations}
    \end{subfigure}
   \caption{Looking at the behavior of several queries (\subref{subfig:distance_per_iterations}), one can see that the general area is found very quickly by the search graph.
   Afterwards, the local neighborhood is explored and shortly after finding the last improvement (indicated by the marker), the query is terminated by the stopping criterion.
   The effectiveness of the stopping criterion also shows in the statistics (\subref{subfig:query_iterations}).
   The distribution of total iterations closely follows the distribution of iterations necessary to make the final improvement, only lagging behind by a handful of iterations.}
   \label{fig:query_behavior}
\end{figure}

It is instructive to look at the behaviour of the query over time.
In Figure~\ref{fig:query_behavior}, the initial distance of the query to the starting point, i.e.\ the closest point on the top layer, is drastically reduced already after very few iterations.
Here, an iteration stands for fetching the best point from the priority queue and calculating the distance to all its neighbors. %
Most time is spent to carefully explore the neighborhood around the true nearest neighbor.
Quickly after locating the best candidates, the query is terminated by the stopping criterion to prevent unnecessary iterations.
This effect does not only show up in the small subset of plotted query distances, but also statistically in the histogram over all queries.

\subsubsection{Out of GPU-memory Queries}
\label{subsec:out_of_memory}
When querying on billion-scale datasets, the amount of necessary GPU-memory will quickly exceed the available memory even on an 8 NVIDIA Tesla V100 system. %
While SIFT1B together with the low-profile graph GGNN$^{8\ddagger}$ still fits on the GPUs, with the GGNN$^{8}$ graph, it exceeds the available memory by 2 shards ($4.2$ GiB). %
On DEEP1B, the dataset alone already surpasses the available memory, and with GGNN$^{8}$ an additional $28$ GiB per GPU need to be loaded per query pass. %
This results in a bound that is defined by the host to device memory throughput ($\approx12$~GiB/s~\cite{jia2018dissecting}),
\ie, for 10k queries, this results in a minimum of 35\si{\micro\second}/query on SIFT1B and 233\si{\micro\second}/query on DEEP1B.
These bounds are shown in Figure~\ref{fig:performance_comparisons}.
Note that this loading time can be hidden completely by performing an overall costlier query, e.g. using more queries, increased $\tau_q$, or larger caches.
For easy comparison with future approaches on devices with sufficient memory, we also report the computation time without the memory transfers.

%% file: tex/conclusion.tex
\section{Conclusion}
\label{sec:conclusion}
Our proposed parallel GPU-based search algorithm is surpassing all state-of-the-art algorithms in terms of speed by a large factor and represents the fastest ANN query technique while maintaining recall rates above 99\%. %
The efficient query algorithm further accelerates the parallel construction and merging of sub-graphs. Our hierarchical construction scheme creates high-quality kNN graphs with graph-diversification links for very efficient traversal. %
It is easily deployed in multi-GPU systems to cope with very large-scale datasets.
It is the first ANN search method capable of constructing effective search structures (R@1 $\geq$ 0.99) for billion-scale datasets in less than an hour.
For million-scale datasets, search-graphs of sufficient quality can often be constructed within seconds,
allowing for quick nearest-neighbor searches also on intermediate data structures as a part of larger GPU-based algorithms, e.g.\ in machine learning applications.

Currently, our method computes the exact distance to all visited high-dimensional points.
As the number of distance calculations dominates the query time, a compressed representation of the data vectors could lead to further acceleration.

%% file: tex/appendix.tex
\section{Hierarchical Construction Example}
\label{sec:construction_example}
\begin{figure}[htb]
 \centering
 \includegraphics[width=.9\columnwidth]{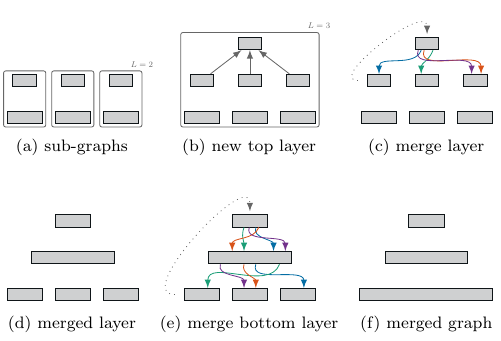}
 \caption{Bottom-up-top-down construction of our graph hierarchy.
 a) Disjoint batches/sub-graphs provide a few sampled points which are fused into a common new top layer b).
 For the top layer, a symmetric kNN-graph is constructed by brute force.
 c) The next finer layer is fused by carrying out hierarchical kNN-searches from the top which will find and link neighborhoods across sub-graphs, d) establishing one consistent kNN-layer.
 This process of e) finding the nearest neighbors for all points in a layer and f) fusing this layer across the sub-graphs is recursively applied until all initial shards are merged.}
 \label{fig:tree_construction}
\end{figure}

To visualize the individual steps in the construction of the hierarchy, Figure~\ref{fig:tree_construction} shows an example with growth rate $g=3$
where layer $1$ has already been constructed by merging 9 bottom layer batches into 3 sub-graphs of height $2$.
To merge the individual sub-graphs further,
first, $s$ points are selected equally from the current top layers of the $g$ sub-graphs, creating a new top layer (\Select). %
Starting from the top, layer by layer,
a hierarchical query for the $k$ nearest neighbors of all points in the current layer is performed
which can access all $g$ sub-graphs via the new top layer.
(On the new top layer, the $k$ nearest neighbors can again be determined efficiently using brute-force.)
The kNN lists within the layer are then updated using the result of these queries (\Merge).
Afterwards, symmetric links are inserted (\Sym). %
Once all layers are processed, the $g$ sub-graphs form one consistent search-graph,
where each point can potentially reach all others. %

\section{Parallelization of Graph-based methods}
\label{sec:appendix_parallelization}
\begin{figure}[ht]
    \centering
    \includegraphics[width=0.8\linewidth]{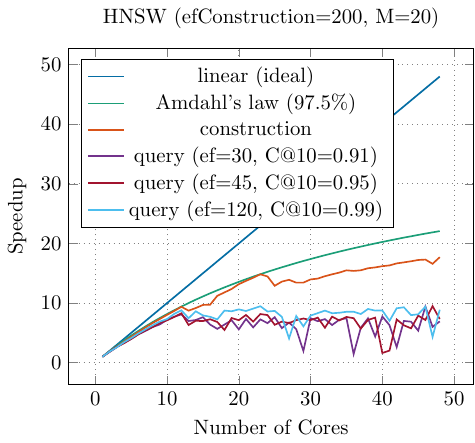}
    \caption{Parallelization of the HNSW algorithm on two Intel(R) Xeon(R) CPU E5-2650 v4 @ 2.20GHz (2x12 cores with 24 threads each).
    First, all 12 physical cores on the first CPU were used, then the 12 cores of the second CPU were added. Finally, the logical cores were added.
    Speedup is computed as the execution time on a single core divided by the parallelized time.
    The trendline using Amdahl's law has been fitted optimistically.}
    \label{fig:HNSW_parallelization}
\end{figure}

Graph-based methods receive impressive recall numbers even for high-dimensional datasets, e.g. HNSW~\cite{malkov2018efficient}. %
Nonetheless, there is a caveat to it. Every comparison between two points needs to load the complete vector from memory. %
Therefore, graph-based methods require a huge memory throughput, that gets especially prominent when considering parallelization. %
In contrast, quantization-based methods are already utilizing the huge advantages in memory throughput offered by GPUs, \eg 900 GB/s NVIDIA Tesla V100 vs. 76.8 GB/s Intel Xeon E5-2650 v4. %
Figure~\ref{fig:HNSW_parallelization} demonstrates the limits of CPU-based parallelization approaches.
Hence, our work focuses on enabling graph-based kNN structures on GPUs, where we do not have the luxury of cheap global or sequential update steps, excessive resources, nor simple dynamic neighborhood lists. %

\begin{table*}[tp]
    \centering
    \caption{Query configurations for the NVIDIA Tesla V100 (CUDA compute capability 7.0).}
    \label{tab:query_config}
    \begin{tabular}{ccccccccc}
        \toprule
        Dataset & \Best & \PrioQ & \Visited & max. iter. & block size & registers & shared memory & occupancy \\
        \midrule
        \multirow{2}{*}{SIFT1M} &  10 & 54 &  384 &  400 &  32 & 56 & 2200$^\ddagger$ / 2232 & 0.5 \\
                                & 100 & 28 &  896 & 1000 &  64 & 48 & 4796 & 0.625 \\
        NyTimes                 &  10 & 22 & 2016 & 2000 & 128 & 51 & 8536$^\ddagger$ / 8760 & 0.5625 \\
        GloVe 200               &  10 & 22 & 2016 & 2000 & 128 & 51 & 8760 & 0.5625  \\
        GIST                    &  10 & 22 & 2016 & 2000 & 128 & 56 & 8740 & 0.5625 \\
        \midrule
        SIFT1B & \multirow{2}{*}{10} & \multirow{2}{*}{54} & \multirow{2}{*}{384} & \multirow{2}{*}{400} & \multirow{2}{*}{32} & \multirow{2}{*}{56} & 2200$^\ddagger$ / 2232 & \multirow{2}{*}{0.5} \\
        DEEP1B & & & & & & & 2200 & \\
        \bottomrule
    \end{tabular}
\end{table*}

\section{Choosing Parameters}
Our method can be tuned for optimal performance using various parameters.
In the following, we provide guidelines on choosing good parameters for the construction and query process.

\subsection{Choosing Search-Graph Construction Parameters}
Our search graph construction process can be tuned by the following parameters:
\par
$k = |\mathcal{N}_x|$ selects the number of neighbors, i.e., outgoing links, per point in the graph and is best kept low to prevent over-exploration during query.
For highly skewed or very high-dimensional datasets, large neighborhoods seem necessary to bridge gaps and achieve high recalls.
\par
$s$ is the segment size in which points are grouped and nearest neighbors are determined in a brute-force fashion to fill the initial neighbor lists.
As a result, $s > k/2$ is a requirement.
Simultaneously, it is also the size of the top-layer which is used as the set of starting points for each query.
Using multiples of 32 makes good use of the GPU's resources which always executes 32 threads simultaneously in a \emph{warp}.
\par
$l$ chooses the number of layers in the hierarchy.
Larger hierarchies can bridge wider gaps in the dataset more easily, but don't help much in the case of individual outliers.
The larger $l$, the longer the construction will take.
Empirically, we found $l=4$ to be the ideal choice for most datasets.
\par
$r$ chooses the number of refinement iterations which repeat the merging and symmetric linking steps performed during construction.
For most datasets, using $r=2$ can significantly boost query performance, while additional refinements do not result in further improvements of query performance.
For the NyTimes dataset, we did not observe any improvement, most likely due to its comparably small size.

\subsubsection{Graph Size}
In some applications, the size of the search graph can be a limiting factor.
Especially for larger datasets, it can become problematic to fit the dataset and the search graph onto the GPU.
In such cases, we resort to conventional sharding.

To determine the precise memory requirements, one must first determine the exact integer growth factor $g$ and the resulting number of points on the upper layers $N_{\mathrm{upper}}$.
There are $N_\mathrm{upper} = s\sum_{i=0}^{l-2} g^i$ points in the upper layers of the hierarchy
where the growth factor $g \approx \sqrt[l-1]{\frac{N}{s}}$ may be rounded up or down such that the bottom-level segment size $s_0 \approx \frac{N}{g^{l-1}}$ is closest to $s$ except to ensure that $s_0 > k$.

To determine the memory required by the search-graph, one can then simply sum up the following:
\begin{itemize}
    \item $(N+N_{\mathrm{upper}}) \cdot k \cdot 4~\mathtt{Bytes}$ for the graph's edge list.
    \item $2 \cdot N_{\mathrm{upper}} \cdot 4~\mathtt{Bytes}$ for translating higher-level indices into the level below and into the base-level indices.
    \item $2 \cdot 4~\mathtt{Bytes}$ for $d_{\mathrm{nn}_1}^+$ and $\bar{d}_{\mathrm{nn}_1}$.
\end{itemize}

As an example, for SIFT1M with $k=24$, the graph takes the following amount of memory: ($g=32$)
\begin{equation*}
    (1033824 \cdot 24 + 2 \cdot 33824 + 2) \cdot 4~\mathtt{B} ~\approx~ 94.9~\mathtt{MiB}
\end{equation*}

During construction, additional temporary memory is required:
\begin{itemize}
    \item $N \cdot k \cdot 4~\mathtt{Bytes}$ for temporary edge lists.
    \item $N \cdot 4~\mathtt{Bytes}$ for random numbers used for selection.
    \item $N \cdot (k_{sym}+1) \cdot 4~\mathtt{Bytes}$ for temporary inverse link lists.
    \item $N \cdot 4~\mathtt{Bytes}$ for the nearest neighbor distance per point.
    \item temporary storage for reduction using CUB
\end{itemize}
As not all of these are used simultaneously, about $N \cdot (k+1) \cdot 4~\mathtt{Bytes}$ suffice in practice,
which is roughly as much as is needed for the graph itself.

\subsection{Choosing Query Parameters}
To make the most use of the GPU's resources, the parameters of the query kernel can be carefully tuned.
The query uses \emph{registers} to cache the query vector for efficient distance computations,
whereas our cache structure resides in \emph{shared memory}.

As discussed in the paper, we can afford to use \Best lists as short as the number of neighbors to be searched for.
However, we always search for at least 10 neighbors. This comes at negligible additional cost for queries for fewer neighbors but keeps the stopping criterion stable.

When aiming for high accuracy, it turned out that large \Visited lists are of utmost importance,
as most time is spent to carefully explore the local neighborhood of a query.
Being able to enumerate each visited point makes sure that the query does not perform any cycles.
We also limit the maximum number of iterations performed by the query such that the visited list is not exceeded by more than a few points at most.

Concerning the \PrioQ, we found that using larger priority queues is only beneficial up to a point and often very short priority queues perform surprisingly well,
allowing us to only cache as little as 32 or 64 distances in our queries.

Finally, the \emph{block size} can be tuned to achieve higher \emph{occupancy} on the GPU when more registers are needed, e.g., for high-dimensional datasets,
or more shared memory, when using a large cache.
The ideal values will vary somewhat depending on the used GPU but small block sizes (e.g., 32) perform better during the reduction-based distance computation, especially for lower-dimensional (e.g., 128) datasets.
For more costly queries, however, larger block sizes may be necessary to achieve higher occupancy which allows to hide latencies when accessing data in the GPU's main memory.

\subsubsection{Used Query Parameters}
The detailed query parameters used for evaluating the performance of our method on each dataset can be found in Table~\ref{tab:query_config}.
This includes the GPU-specific register and shared memory usage, as well as the resulting occupation on the device.

We use the same parameters across the more complicated datasets.
For SIFT1M, a cheaper query suffices.
The query for GIST needs a few more registers to store the high-dimensional query vector.
The queries for NyTimes and GloVe 200 have some shared memory overhead due to the cosine-similarity distance metric.
There is almost no overhead due to the dataset size, so even the queries on SIFT1B and DEEP1B can use the same configuration as the query on SIFT1M with the same resulting register and shared memory usage.
(There is a minor overhead when dealing with billion-scale datasets due to the necessity of 64-bit indices when addressing base-vectors or neighbors in the graph.)
In all cases, our query reaches an occupancy of at least 50\%, which allows to effectively hide memory-access latencies.

\section{Performance on Different GPUs}
\begin{figure}[ht]
    \centering
    \includegraphics{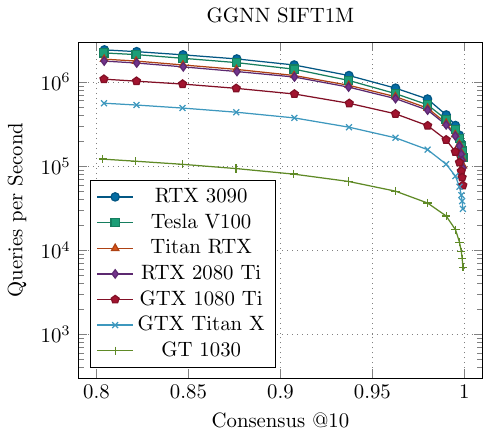}
    \caption{Performance comparison across different GPUs.}
    \label{fig:gpu_comparison}
\end{figure}

In the main paper, we only show results using the NVIDIA Tesla V100, which is a high-end datacenter GPU.
Figure \ref{fig:gpu_comparison} demonstrates the performance of our method on a wide range of GPUs,
including the low-end GT 1030.
All GPUs used the same query configuration shown in Table~\ref{tab:query_config}.
On the GT 1030 and GTX 1080 Ti, increasing the blocksize to 64 resulted in slightly faster queries.

On all GPUs, our method performs equally well in terms of quality but the number of queries per second depends on the memory bandwidth and computational power that the GPU is able to provide.
E.g., with about 5\% of the V100's memory bandwidth, the GT 1030 is also only able to provide about 5\% of the query performance.
The construction time is similarly impacted, as shown in Table~\ref{tab:sift1m_gpu_construction_times}, which also shows the memory bandwidth of the compared GPUs.

Finally, the consistent query quality across the GPUs also demonstrates the stability of the search-graph construction process, despite its largely independent construction and random selection steps.
While the shown results are each based on different, newly constructed search-graphs, their consensus at the fixed slack factor ($\tau_q$) intervals is almost identical.

\begin{table}[ht]
    \centering
    \caption{Construction times for the GGNN search-graph for SIFT1M on various GPUs.}
    \begin{tabular}{lrrr}
      \toprule
      GPU         & Memory Bandwidth & Construction Time \\
      \midrule
      RTX 3090    &                   936 GB/s &  21.7 s \\
      Tesla V100  &                   900 GB/s &  21.8 s \\
      Titan RTX   &                   672 GB/s &  26.7 s \\
      RTX 2080 Ti &                   616 GB/s &  29.3 s \\
      GTX 1080 Ti &                   484 GB/s &  59.1 s \\
      GTX Titan X &                   336 GB/s & 102.1 s \\
      GT 1030     &                    48 GB/s & 586.8 s \\
      \bottomrule
    \end{tabular}
    \label{tab:sift1m_gpu_construction_times}
\end{table}

\section{Query Starting Points}
Every graph based search needs to start on at least one point in the graph. %
Our algorithm uses all the points at the top layer as \textbf{starting points}. %
These points are well connected in the graph due to the construction design, which is of particular interest during construction itself. %
However, In a well converged flat search-graph, the actual starting points are not that important in terms of quality. %
The option to start from the \textbf{centroid} point is proposed by Li~\etal \cite{li2019approximate}. %
We conducted an experiment to compare the different methods. %
Additionally, we also test the performance when a search starts at the furthest point from the centroid, which is denoted as \textbf{worst}. %
The centroid or worst point is cheaply determined during construction with no measurable overhead.
The experiment is performed on a lower spec NVIDIA GTX Titan X GPU, compared to the main paper. %
Table~\ref{tab:startingpoints} shows the results for the 1, 2, and 5 million point subsets of the SIFT1B dataset. %
While c@10 is on par for all query start point options, we see some improvements in terms of speedup when using our \textbf{starting points}. %

\begin{table}[ht]
    \centering
    \caption{Comparison of query start options for different slack factors $\tau$ and subsets of the SIFT1B. Execution times are evaluated on an NVIDIA GTX Titan X GPU.}
    \resizebox{\linewidth}{!}{%
    \begin{tabular}{c|cc|cc|cc|c}
      \toprule
               \multirow{2}{*}{$\tau$}& \multicolumn{2}{c|}{starting points} & \multicolumn{2}{c|}{centroid point} & \multicolumn{2}{c|}{worst point} & \multirow{2}{*}{\shortstack{SIFT1B \\ subset}}\\
         & c@10 & us/query & c@10 & us/query & c@10 & us/query &  \\
  \midrule
  \multirow{3}{*}{0.6} & 0.983 & 7.4 & 0.983 & 8.5 & 0.983 & 8.5 & 1M \\
   & 0.978 & 8.1 & 0.978 & 9.3 & 0.978 & 9.3 & 2M \\
   & 0.969 & 8.8 & 0.968 & 10.3 & 0.969 & 10.3 & 5M \\
  \midrule
  \multirow{3}{*}{0.7}
   & 0.993 & 12.2 & 0.994 & 13.7 & 0.993 & 13.7 & 1M \\
   & 0.990 & 13.2 & 0.990 & 14.9 & 0.990 & 14.8 & 2M \\
   & 0.985 & 14.0 & 0.984 & 16.1 & 0.985 & 17.0 & 5M \\
  \bottomrule
    \end{tabular}}
    \label{tab:startingpoints}
\end{table}

\section{Impact of Partitioning}
In our experiments, we are using all datasets as they are, i.e. without any argumentation or pre-analysis. %
All points in the dataset are read in as they are laid out in the files (\emph{constant}). %
Our algorithm splits data into multiple different partitions throughout the construction process. %
When using multi-GPU variants, further partitions are introduced by sharding. %
Table~\ref{tab:shuffle} demonstrates the effect of constructing the search-graph on a randomly \emph{shuffled} dataset to mix up the points grouped together in each otherwise deterministic partition. %
The experiment is conducted on on the 1 million subset of SIFT1B with 4 shards. %
The mean $\mu$ and standard deviation $\sigma$ over the consensus@10 are computed across 100 iterations. %
While $\sigma$ is increased in the \emph{shuffled} case, $\mu$ is on par with the constant read in, which is expected due to our randomized selection process for upper-level points. %

\begin{table}[ht]
    \centering
    \caption{Analysis of partitioning impact: Mean $\mu$ and standard deviation $\sigma$ are computed on the 1M subset of SIFT1B with 4 shards over 100 iterations.}
    \begin{tabular}{c|cc|c}
      \toprule
      \multirow{2}{*}{$\tau$}& \multicolumn{2}{c|}{c@10} & \\
          & $\mu$ & $\sigma$ & \\
      \midrule
      \multirow{2}{*}{0.6}
      & 0.9956 & 1.8e-5 & constant \\
      & 0.9959 & 2.0e-4 & shuffled \\
      \midrule
      \multirow{2}{*}{0.7}
      & 0.9989 & 1.4e-5 & constant \\
      & 0.9988 & 9.5e-5 & shuffled \\
  \bottomrule
    \end{tabular}
    \label{tab:shuffle}
\end{table}

\section{Impact of the Symmetric Linking Constraint and Refinement}
To analyze the impact of our proposed symmetric linking constraint introduced in Section \ref{subsubsec:sym_link_constraint}
and search-graph refinement introduced in Section \ref{subsec:graph_refinement},
we inspect the query performance on SIFT1M for several configurations.

\begin{figure}[ht]
 \centering
 \includegraphics{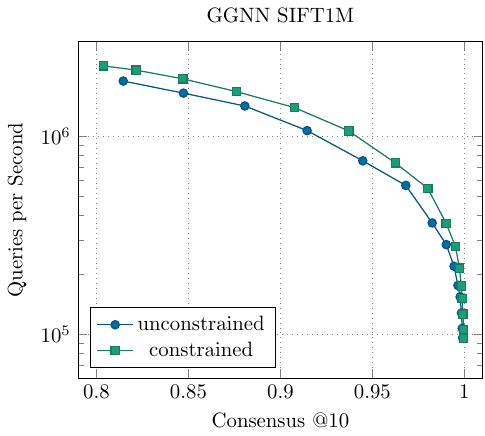}
 \caption{Comparing query performance using symmetric linking without and with the additional constraint introduced in Section \ref{subsubsec:sym_link_constraint}.}
 \label{fig:sym_constraint}
\end{figure}

We inspect query performance without and with the additional symmetric linking constraint introduced in section \ref{subsubsec:sym_link_constraint} in Figure~\ref{fig:sym_constraint}.
Search-graph construction took $21.2$ seconds without the constraint and $22.1$ seconds with the constraint.
Without the additional constraint, symmetric linking took about $250$ ms more, but merging took about $1.1$ seconds less construction time.

These results confirm that constraining the ``symmetric'' query for reverse paths from known nearest neighbors
results in the addition of better links for graph diversification.
With the additional constraint in place,
following queries explore a richer set of candidate points, which takes slightly more time to explore but also results in increased accuracy.
This is part of what allows our method to achieve (close to) perfect recall rates even on billion-scale datasets.
At the same time, more queries can be processed per second at the same accuracy.

\begin{figure}[ht]
 \centering
 \includegraphics{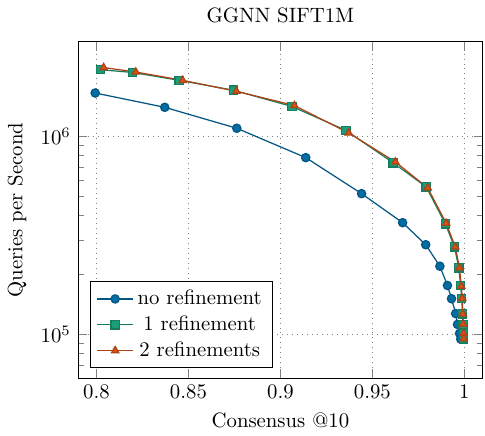}
 \caption{Query performance at different levels of refinement.}
 \label{fig:refinement}
\end{figure}

When inspecting the query performance at different levels of refinement as shown in Figure~\ref{fig:refinement},
we see that query performance is increased significantly after the first refinement iteration.
Refining the search-graph a second time can further boost query performance by a small amount in terms of both speed and accuracy.
Further refinements result in no noticeable changes in query performance.
Therefore, we generally perform two refinement iterations during search-graph construction to achieve optimal query performance while keeping construction time to a minimum.